\documentclass[10pt,journal,compsoc]{IEEEtran}
\usepackage{ragged2e} 
\usepackage{amsmath,amsfonts}
\usepackage{array}
\usepackage{graphicx}
\usepackage[caption=false,font=footnotesize,labelfont=rm,textfont=rm]{subfig}
\usepackage{textcomp}
\usepackage{stfloats}
\usepackage[table,xcdraw]{xcolor}
\usepackage{url}
\usepackage{multirow}
\usepackage{booktabs}
\usepackage{verbatim}
\usepackage[ruled,linesnumbered]{algorithm2e}
\usepackage{bbding}
\usepackage{stfloats}
\usepackage[accsupp]{axessibility}
\usepackage{algpseudocode}

\usepackage{todonotes}

\ifCLASSOPTIONcompsoc
  \usepackage[nocompress]{cite}
\else
  \usepackage{cite}
\fi
\usepackage[pagebackref,breaklinks,colorlinks]{hyperref}
\begin{document}

\title{Unsupervised Pre-training with Language-Vision Prompts for Low-Data Instance Segmentation}

\author{Dingwen Zhang, Hao Li, Diqi He, Nian Liu, Lechao Cheng, \\ Jingdong Wang, \textit{IEEE Fellow}, Junwei Han, \textit{IEEE Fellow}
        
\IEEEcompsocitemizethanks{
\IEEEcompsocthanksitem Dingwen Zhang, Hao Li, Diqi He and Junwei Han are with the Brain and Artificial Intelligence Lab, Northwestern Polytechnical University, Xi'an, Shaanxi 710000, China.
\IEEEcompsocthanksitem Nian Liu is with the Inception Institute of Artificial Intelligence. 
\IEEEcompsocthanksitem Lechao Cheng is with the School of Computer Science and Information Engineering, Hefei University of Technology, Hefei, 230601, China. E-mail: chenglc@hfut.edu.cn. 
\IEEEcompsocthanksitem Jingdong Wang is with Department of Computer Vision, Baidu Inc. E-mail: wangjingdong@baidu.com.
}
\thanks{This work was supported in part by the National Natural Science Foundation of China under Grant 62293543, Grant U21B2048 and Grant 62106235.}
}


\IEEEtitleabstractindextext{%
\justify
\begin{abstract}
    In recent times, following the paradigm of DETR (DEtection TRansformer), query-based end-to-end instance segmentation (QEIS) methods have exhibited superior performance compared to CNN-based models, particularly when trained on large-scale datasets. Nevertheless, the effectiveness of these QEIS methods diminishes significantly when confronted with limited training data. This limitation arises from their reliance on substantial data volumes to effectively train the pivotal queries/kernels that are essential for acquiring localization and shape priors.
    To address this problem, we propose a novel method for unsupervised pre-training in low-data regimes. Inspired by the recently successful prompting technique, we introduce a new method, \emph{Unsupervised Pre-training with Language-Vision Prompts (UPLVP)}, which improves QEIS models' instance segmentation by bringing language-vision prompts to queries/kernels.
    Our method consists of three parts: 
    (1) \emph{Masks Proposal}: Utilizes language-vision models to generate pseudo masks based on unlabeled images. 
    (2) \emph{Prompt-Kernel Matching}: Converts pseudo masks into prompts and injects the best-matched localization and shape features to their corresponding kernels. 
    (3) \emph{Kernel Supervision}: Formulates supervision for pre-training at the kernel level to ensure robust learning. 
    With the help of our pre-training method, QEIS models can converge faster and perform better than CNN-based models in low-data regimes. 
    Experimental evaluations conducted on MS COCO, Cityscapes, and CTW1500 datasets indicate that the QEIS models' performance can be significantly improved when pre-trained with our method.
    Code will be available at: \href{https://github.com/lifuguan/UPLVP}{https://github.com/lifuguan/UPLVP}.
\end{abstract}

}

\maketitle

\section{Introduction}
\label{sec:introduction}

\begin{figure}[t]
  \centering
  \includegraphics[width=1\linewidth]{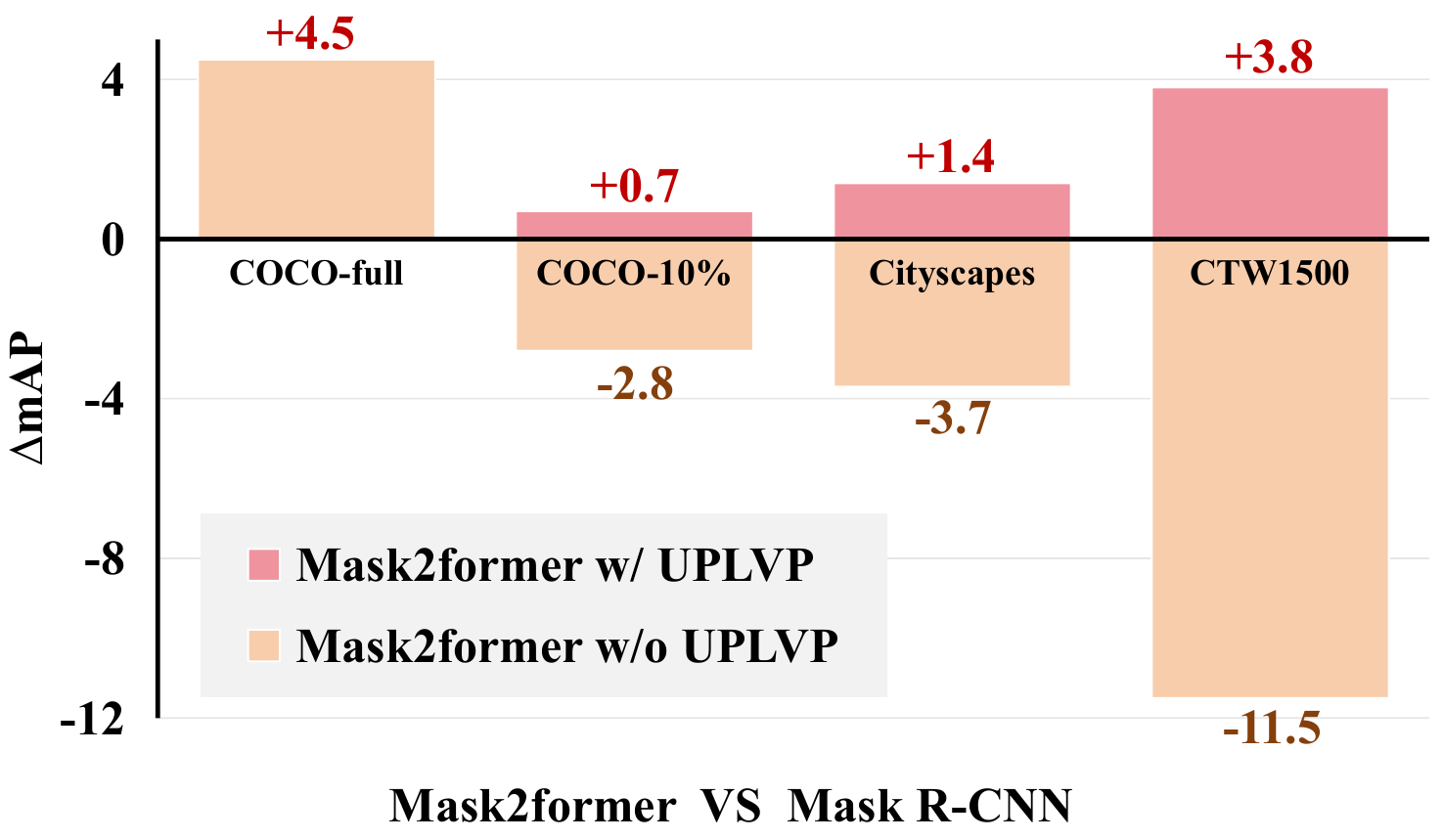}
  \caption{Performance of Mask2former in different methods using Mask R-CNN as the benchmark. Here ConvNeXt-V2 is used as the backbone of Mask R-CNN. Mask2former is able to outperform Mask R-CNN with large-scale datasets (COCO-full), but it cannot perform as well as Mask R-CNN with small datasets (COCO-10\%, Cityscapes and CTW1500) since it's hard to learn localization and shape priors. Our proposed unsupervised pre-training method based on language-vision prompts not only boosts the vanilla Mask2former significantly but also helps to achieve comparable performance compared with Mask R-CNN.}
  \label{fig:intro}
\end{figure}

    \IEEEPARstart{I}{nspired} by the state-of-the-art model DETR~\cite{detr} used for object detection, many innovative Query-based End-to-end Instance Segmentation (QEIS) models~\cite{queryinst,mask2former,maskformer,maxdeeplab,istr,knet} have been developed to address instance segmentation task in unique ways.
    In contrast, modern CNN models perform the instance segmentation task indirectly by formulating the localization problem in a large set of proposals~\cite{maskrcnn}, window centers~\cite{tensormask, instanceFCN}, or location-based masks~\cite{fcos,solov1,solov2}.
    A typical example is Mask-RCNN~\cite{maskrcnn}, which employs a well-structured region proposal network(RPN) to generate candidate bounding boxes and subsequently refine the proposals by filtering out those of poor quality. 
    Although the current approach simplifies the learning process for object localization, manual design of non-maximum suppression (NMS) remains essential to discard redundant predictions and preserve those with higher Intersection over Union (IoU) values with ground truth labels. Unlike CNN-based techniques that rely on a large number of proposals for local region features extraction and prediction, QEIS models use dynamic kernels/queries to gain an understanding of object properties automatically, encompassing diverse location and shape information.
    The Hungarian match algorithm serves as a useful tool for crafting effective kernels/queries, thereby obviating the need for redundant hand-crafted anchors and post-processing methods such as NMS. This is attributed to the kernels/queries's capacity to learn more precise location and shape priors. However, owing to the dynamic nature of kernels or queries, they must learn the general spatial distribution and shapes of objects in a data-driven manner to adapt to various input images. In scenarios characterized by a scarcity of annotated data during the fine-tuning phrase, known as low-data regimes~\cite{detreg}, QEIS models often exhibit more pronounced declines in performance compared to CNN-based methods, as depicted in Figure~\ref{fig:intro}. To illustrate this, we compare Mask2Former~\cite{mask2former} with Mask R-CNN, the former being a representative QEIS model for universal segmentation and the latter a typical CNN-based model. Here we use ConvNeXt-V2~\cite{convnextv2} as the backbone of Mask R-CNN due to its outstanding performance. 
    It can be seen that Mask2former can outperform Mask R-CNN when trained with the entire COCO dataset, but its performance drops significantly with small-scale datasets (COCO-10\%, Cityscapes, and CTW1500).
    
    It is worth noting that the QEIS models still possess significant potential. Equipped with the ability to acquire ample and high-quality localization and shape priors, they are capable of performing on par with, or even surpass, CNN-based methods. This prompts us to explore strategies to facilitate QEIS models to quickly grasp localization and shape priors, particularly in low-data situations.
    Adopting unsupervised pre-training is a promising solution to train models before the actual training phase. This method does not require any modifications to the off-the-shelf model architectures or extra training data. However, most existing unsupervised pre-training methods~\cite{detreg,updetr,swav,densecl,mocov2} only focus on the backbone, neglecting the instance segmentation prediction heads. This poses a challenge, as the prediction heads encode crucial localization and shape priors, with kernels/queries playing a crucial role in this process.
    In the field of object detection, some works~\cite{updetr,detreg} do pre-train their complete model architectures, albeit employing pseudo bounding boxes for training. However, many of those pseudo bounding boxes lack object content and consequently fail to generate pseudo instance masks for instance segmentation.
    The pioneering work that specifically designs for the instance segmentation is FreeSOLO~\cite{wang2022freesolo}. This method mainly focuses on generating pseudo masks and uses them to directly supervise the model during training. Nevertheless, this approach still learns the localization and shape priors of objects in a data-driven way. Consequently, the generation of high-quality pseudo masks entails laborious steps.
    
    
    To alleviate the limitations of low-data QEIS models, we propose a new method termed \emph{Unsupervised Pre-training with Language-Vision Prompt (UPLVP)}. Inspired by the success of the prompting mechanism in visual tasks~\cite{bert,clip,coop,cooop,vpt,chen2024virtual}, our method incorporate language-vision prompts directly into the kernels/queries of the model to improve localization and shape priors. 
    First, we derive pseudo masks indicating potential location and shape information of instances. These masks serve as the basis for generating prompting, which are then integrated into kernels/queries to imbue them with location and shape knowledge. This approach can speed up the model convergence during training and improve its ability to learn with low amounts of data.
    
    Specifically, given an unlabeled image, pseudo masks corresponding to the image are generated through a \emph{Pseudo Masks Generation} method. Notably, approaches such as FreeSOLO~\cite{wang2022freesolo} and UPSP~\cite{upsp} adopt a saliency mechanism that focuses solely on distinguishing foreground and background instances, assigning binary category labels (0 or 1) accordingly. While this simplifies pseudo mask generation, it overlooks the incorporation of category information.
    For image segmentation task that involves visual semantic information and localization, it is essential to incorporate category information. To this end, we leverage large language-vision models such as CLIP~\cite{clip} and OpenSeg~\cite{openseg} to harness semantic understanding across diverse categories. 
    After obtaining the masks, previous methods use them as pseudo labels to supervise the model's predictions. 
    However, there is still valuable information that can be leveraged, such as region location. Instead of learning directly from the noisy pseudo masks, we use them to generate corresponding regional features and obtain language-vision prompts from them before the training phase.
    Subsequently, a \emph{Prompt-Kernel Matching} module is responsible for matching prompts with kernels. This module uses location and shape information from the prompts to modify the kernels accordingly, which are then used by the QEIS decoders.
    What's more, an \emph{Auxiliary Kernel Supervision} scheme is also introduced to supervise the model learning to maintain the robustness of kernels. 
    See Figure~\ref{fig:overview} for overview.

    In the experiments, our proposed UPLVP outperforms all other existing algorithms which focus on pre-training backbones 
    in low-data regimes on four datasets, verifying that our method can address the shortcomings of previous work.
    Our method can be used for most QEIS models as a plug-and-play pre-training step which can greatly speed up convergence and get better performance without any increase in parameters or memory. 
    In detail, our method achieves two \emph{desiderata} on downstream tasks:
    (1) making QEIS models reach the same convergence speed as CNN-based methods and
    (2) making QEIS models obtain comparable or even better performance than CNN-based methods on most downstream datasets.
    In the ablation studies we conduct, we discover that our method presents a big tolerance to the quality of pseudo masks. Thus, we can easily improve performance without a sophisticated and time-consuming pseudo mask generation method, as required in FreeSOLO~\cite{wang2022freesolo}.
    
    Our contributions can be summarized as follows:
\begin{itemize}
\item By proposing a novel unsupervised pre-training approach, we made an early effort to address the problem that QEIS models require a large amount of data for training. 
\item To facilitate the desired unsupervised pre-training, we propose to generate high-quality pseudo masks, by the first time, utilizing two Language-Vision Models and a newly designed prompt injecting mechanism to introduce localization and shape priors into kernels of the QEIS model. 
\item Experimental results on the widely-used MS COCO, Cityscapes, and CTW1500 datasets demonstrate that our approach can speed up the convergence of the QEIS model and obtain significant performance improvement in low-data regimes. 
\end{itemize}

    This work is an extension of~\cite{upsp}. Compared with the conference version, we extend our study in the following aspects: 1) Instead of simply adopting the saliency mechanism, we establish a new framework by introducing the language-vision models to generate pseudo masks, which can improve the quality of the obtained mask proposals as well as incorporate additional category information to those proposals. 2) We design a simple yet effective loss, called auxiliary loss, for the classification head to prevent over-fitting of the features. 3) We conduct more comprehensive experiments to explore the effectiveness of the newly proposed modules and loss and demonstrate the superior capacity of the new learning framework. 4) We additionally perform experiments on a downstream task, few-shot instance segmentation, demonstrating that our approach has the potential to be applied to a wide range of downstream tasks.

\section{Related Work}
\subsection{Query-Based End-to-End Instance Segmentation}
    As the Transformer successfully comes into the world and is deployed in vision tasks, DETR~\cite{detr} has proposed a novel object detector based on object queries. 
    The detection task has been split into a series of iterative prediction problems making DETR the first end-to-end model eliminating anchors or NMS to remove redundant prediction boxes. 
    Under this kind of pattern, each query is responsible for attending to a specific local region similar to anchors in CNN-based models. The difference is queries/kernels in QEIS are learned during training iteratively to learn more precise and diverse localization and shape priors. With the help of the Hungarian match algorithm where each ground truth label is only assigned to one prediction, QEIS models can remove NMS.
    Instead of proposing dense instance proposals in the prediction head,  queries/tokens/kernels are used to be responsible for individually pooling instance features on the global scale, hence are more flexible and enable to realize end-to-end learning. 
    
    After DETR, many works~\cite{queryinst,mask2former,maskformer,maxdeeplab,istr,knet} follow this paradigm with various improvements to complete the instance segmentation task, which we denote as Query-Based End-to-End Instance Segmentation (QEIS) methods. 
    Gaining inspiration from the idea of SOLO-v2~\cite{solov2}, K-Net~\cite{knet} adopts kernels that convolve with feature maps to predict masks directly. 
    This kernel-mask paradigm enables K-Net to achieve both semantic and instance segmentation tasks consistently by a set of dynamic learnable kernels.
    Building upon Sparse-RCNN~\cite{sparse}, QueryInst~\cite{queryinst} adopts parallel supervision on dynamic mask heads.
    By using multi-scale feature fusion and masked-cross-attention, Mask2Former~\cite{mask2former} improves the accuracy and efficiency of the prediction head.
    In our study, we use K-Net as a typical example of QEIS models and deploy our unsupervised pre-training method on it. Note that our UPLVP method can also be applied to other QEIS-style models freely and improve their performance in low-data regimes, and we prove the fact in the next experiments section~\ref{experiments}. 
 
\begin{figure*}[ht]
  \centering
  \includegraphics[width=0.95\linewidth]{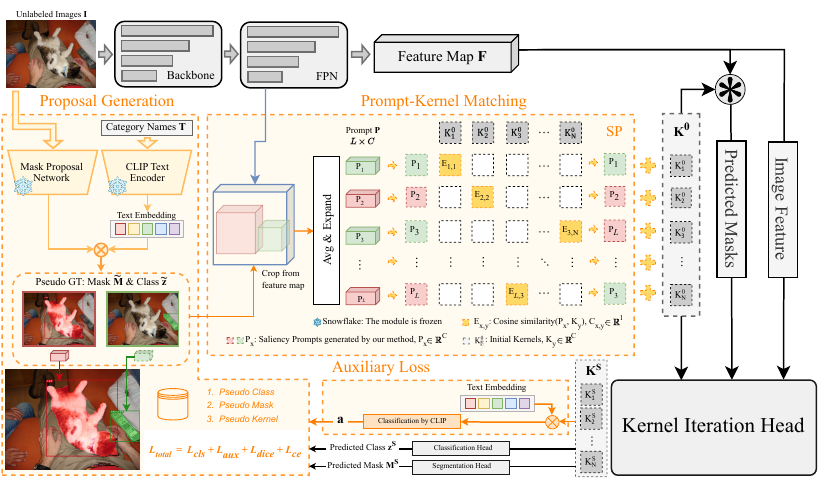}
  \caption{Pipeline of our proposed unsupervised pre-training framework UPLVP. It mainly contains three modules: proposal generation, prompt kernel matching and auxiliary supervision. Here we leverage CLIP and OpenSeg to extract text and vision features respectively. As can be seen, our method is parameter-free.  
  Parts in {\color[HTML]{F56B00} orange} color denote our pre-training method with the corresponding supervision.
  Parts in {\color[HTML]{CCCCCC} Gray} color denote a vanilla QEIS model, where we take K-Net as an example which contains backbone and kernel update iteration head.}
  \label{fig:overview}
\end{figure*}

\subsection{Unsupervised Pre-training}
    Unsupervised pre-training aims to pre-train deep models with carefully designed pretext tasks for boosting the model performance in downstream tasks. 
    Traditional supervised learning methods require a large amount of annotated data as training samples, yet this kind of data is often difficult to obtain and the annotation process consumes time and resources. The emergence of unsupervised pre-training models fills this gap by utilizing unlabeled data for pre-training, providing better initialization parameters and semantic representation for subsequent downstream tasks. Under this pattern, the model is pre-trained in a large-scale dataset followed by fine-tuning in downstream tasks with low training data. 
    
    Most existing unsupervised pre-training methods such as MoCo-v2~\cite{mocov2}, SwAV~\cite{swav}, and DenseCL~\cite{densecl} pre-train their backbones only and neglect the prediction head, thus remaining the data-hungry issue unsolved of QEIS models where queries need to be highly trained. 
    Although models have come into the world like UP-DETR~\cite{updetr} and DETReg~\cite{detreg} which formulate end-to-end unsupervised learning frameworks based on DETR's query-object mechanism creatively, these methods are not workable for instance segmentation task since their pseudo labels only contain backgrounds that can not bring localization and shape knowledge. 
    Specifically, UP-DETR~\cite{updetr} randomly crops image patches from images as pseudo labels. DETReg~\cite{detreg} uses region proposals generated by ResNet and SwAV as pseudo labels.
    Such pseudo labels can severely mislead the segmentation models during the training period.
    Work like FreeSOLO~\cite{wang2022freesolo} concentrates on obtaining high-quality pseudo mask labels getting inspiration from SOLOv2~\cite{solov2}. For the first time, pseudo labels can be utilized to train the instance segmentation models. 
    However, there is also a shortcoming that multiple steps are required in FreeSOLO like pre-training and self-training, just to obtain pseudo masks. How to use the pseudo masks as pseudo labels to supervise the model training is the major issue FreeSOLO needs to solve but it ignores to explore how to use them more efficiently.

    In our study, we follow the pattern of unsupervised pre-training similarly to solve the problem of lacking abundant training data. Pre-training on COCO and fine-tuning on other datasets are conducted successively and only little data is sent to the model during the fine-tuning phase.
    Our experiments indicate that the object localization and shape priors explicitly injected into the kernels can make further improvements compared to solely using pseudo labels as the supervision.

    In the field of deep learning, it is common to come across a vast number of unlabeled samples. However, obtaining labeled samples is a time-consuming and costly process, which leads to a limited number of labeled samples in comparison to the abundance of unlabeled samples. To overcome this issue, people use semi-supervised learning, which involves the incorporation of a large number of unlabeled samples with a small number of labeled samples for training.
    It's important to note that there are some significant differences between our approach and the current popular semi-supervised methods~\cite{nosiy-semi-superviesd}. Our method is self-supervised, meaning the model first undergoes pre-training on a large dataset, followed by fine-tuning on downstream tasks with low data. Additionally, the domains of the downstream tasks are not constrained and may differ from the domains of the pre-training data. However, semi-supervised requires all training data to be in a single domain, or else the model cannot converge, according to our experiments. Therefore, our approach is incomparable to semi-supervised works.

\subsection{Prompting}
    The prompting mechanism technique originates from NLP~\cite{bert} and is continuously noticed and studied. The method is soon transferred to the domain of multi-modal~\cite{clip,coop,cooop} including vision tasks. 
    In the mechanism of prompting, downstream tasks are formulated as a "fill-in-the-blank" issue, like "A photo of a \{object\}" that is used in CLIP~\cite{clip}.
    Here "A photo of a" represents the prompt templates that instruct the language models to elicit useful information from the pre-trained models, and "\{object\}" is the blank word or phrase that models are commanded to predict.
    
    Since the prompting mechanism has a great effect on the "pre-training + fine-tuning" pipeline, it has been receiving great attention in recent days.
    CoOp~\cite{coop} replaces the manually defined language prompts with a series of learnable vectors. 
    Building upon CoOp, CoCoOp~\cite{cooop} is further proposed to produce input-conditional prompts for each image and combine them with existing language dynamic prompts.
    In a word, the prompting mechanism presents its good performance in the language domain and especially vision-language cross domains.
    Recently, VPT~\cite{vpt} is the first to integrate prompting into pure vision tasks, 
    which adds several learnable prompts to the patch tokens of the frozen ViT~\cite{vit} model to fit it for a variety of downstream tasks and datasets without the requirement of fine-tuning the whole ViT model.

    We discover that prompts are utilized in most previous works to get better performance of pre-trained models in downstream tasks. In other words, they only utilize prompts in the fine-tuning phase instead of the pre-training phase. Contrary to their method, we ingeniously tailor the prompting mechanism for our pre-training stage. In this work, box region features are treated as visual prompts that are sent to the model, and segmentation results corresponding to prompts are seen as predictions we want.
    Note that the language-vision prompts we use only work during the pre-training period to instruct the QEIS models to formulate a better prediction head and then are removed when fine-tuning the models.

\subsection{Large-Scale Language-Vision Models}
    The mainstream method for multi-modal large models of language-vision is to use a graphic and text features alignment module, allowing the model to understand both text and image features. Previous works based on Transformer~\cite{transformer} such as~\cite{clip,flava,blip,blip2,bridgetower} generate joint image-text embedding, which can be used for downstream tasks such as zero sample image classification. In our work, we take advantage of powerful pre-trained models CLIP and OpenSeg to generate masks and classification labels as ground truth annotations. These pseudo annotations are then transformed into prompts, which subsequently will be injected into initial kernels.
    
    The basic principle of CLIP~\cite{clip} is to align texts and images in the feature domain. Firstly feature extraction is performed on text and image separately. The backbone of image feature extraction can be either the ResNet or the ViT~\cite{vit} while text feature extraction generally uses the Bert model. Then text and image features are multiplied to calculate cosine distance, whose results of the corresponding different pairs tend to approach 1/0.
    BLIP~\cite{blip} includes two single modal encoders, an image-grounded text encoder, and an image-grounded text decoder. Three types of loss are used: Image Text Contrast Loss which aligns the latent feature space of images and texts, Image Text Matching Loss which models the correlation of multi-modal information between images and texts, Language Modeling Loss which trains the model to generate target captions.
    BLIP-2~\cite{blip2} consists of a pre-trained Image Encoder, a pre-trained Large Language Model, and a lightweight learnable Q-Transformer. In the two consecutive stages, the Q-Transformer is connected to the frozen Image Encoder and the frozen Language-Vision Models respectively to improve multi-modal effects and reduce training costs.

    The large language-vision models exhibit strong zero sample generalization performance on different downstream visual tasks through simple natural language descriptions and prompts. For instance, open-vocabulary segmentation models~\cite{openseg,maskclip,masqclip,simbase,opsnet} address the challenge that closed-set class labels during training and evaluating limit their ability to segment novel objects, by generally leveraging a language-vision model as a mask classifier. Many studies have proved the feasibility of this method since these models integrate dense features of both language and vision. In this task, we adopted a similar approach to generating pseudo labels given the visual features and class prompts for each image without any previous annotations, which can be used in the next training phase.

\section{Methodology}
\label{sec:method}
    In this section, we will discuss our proposed unsupervised pre-training method in detail, using K-Net as an example QEIS model. First, we will provide a brief overview of our UPLVP method, followed by a complete explanation of the UPLVP pipeline and how to use our proposed language-vision prompt to pre-train K-Net.

\begin{algorithm}[ht]
    \caption{Overview of UPLVP during the pre-training phase.}
    \label{alg:alg1}
    \BlankLine
    \KwIn{Input image $\mathbf{I}$; Class names $\mathbf{T}$; Number of update iteration stages $S$; 
    Linear transformation $\phi_j, j=\{1,2\}$; Fully connected layer $\psi_k, k=\{1,2,3,4\}$; 
    Pre-trained weights of CLIP and OpenSeg, $\theta_c$ and $\theta_o$}
    \KwOut{Mask prediction $\mathbf{M}^S$; 
    Predicted categories $\mathbf{z}^S$.}
    \BlankLine
    Initialize $\mathbf{K}^0$ randomly\;
    $\mathbf{F} \leftarrow Backbone(\mathbf{I})$\; 
    $\mathbf{\widetilde{M}}, \mathbf{\widetilde{z}} \leftarrow ProposalGeneration(\mathbf{I}, \mathbf{T}, \theta_c, \theta_o)$\; 
    $\mathbf{P} \leftarrow FeatureToPrompt(\mathbf{F}, \mathbf{\widetilde{M}})$\;
    $index \leftarrow PromptKernelMatching(\mathbf{K}^0, \mathbf{P})$\;
    $\mathbf{K}^0 \leftarrow \mathbf{K}^0 + \mathbf{P}_{index}$\;
    $\mathbf{M}^0 \leftarrow \mathbf{K}^0 * \mathbf{F}$\;
    \While{$i<S$}{
        \textbf{Kernel Iteration}\\
        \qquad $\mathbf{K}^i, \mathbf{M}^i, \mathbf{z}^i=\mathcal{F}^i(\mathbf{M}^{i-1}, \mathbf{K}^{i-1}, \mathbf{F})$\;
        $\mathcal{L} \leftarrow LossFunction(\mathbf{\widetilde{M}}, \mathbf{M}^i, \mathbf{\widetilde{z}}, \mathbf{z}^i)$\;
        $i \leftarrow i+1$\;
    }
    $Output \leftarrow \mathbf{M}^S, \mathbf{z}^S$\;
    \BlankLine
\end{algorithm}

\subsection{Overview of UPLVP}
\label{subsec:knet}
    In the task of instance segmentation, we propose a novel unsupervised pre-training method for QEIS models, termed UPLVP.
    In most instance segmentation scenarios, the number of instances to segment is unknown and varies in different images (average 7.7 in COCO). 
	In QEIS models, the instance-level information is dynamically encoded into $N$ kernels\footnote{$N$ is defined to be larger than the maximum instance number in images.}, each of which needs to focus on a specific location and is responsible for grouping the pixels belonging to instances located in the region. In other words, every kernel is utilized to group pixels that contain similar semantic and localized knowledge.

    As shown in Figure~\ref{fig:overview}, UPLVP mainly contains three parts: Mask Proposal Generation, Prompt-Kernel Matching and Auxiliary Loss. The parts in grey presented in the figure denote a vanilla QEIS model, where we take K-Net as an example.
    Given an input unlabeled image $\mathbf{I}$ and the class names of instances $\mathbf{T}$,
    the mask proposal generation module is responsible for generating pseudo masks $\mathbf{\widetilde{M}}$ and the pseudo categories $\mathbf{\widetilde{z}}$ by leveraging two language-vision models, OpenSeg~\cite{openseg} and CLIP~\cite{clip}.
    Meanwhile, the feature maps $\mathbf{F}$ of the input image are extracted by a backbone and FPN.
    The pseudo masks and the feature maps are then used to extract localization and shape knowledge which are transferred to language-vision prompts $\mathbf{P}$ in the prompt-kernel matching module, which is required to find the best-matched prompt for each kernel and inject it into the initial kernel $\mathbf{K}^0$.

    For kernel iteration in K-Net, the instance segmentation masks $\mathbf{M}$ are gained by conducting convolution on $\mathbf{F}$ with $\mathbf{K}$, and a dynamic kernel update iteration head is designed which takes the segmented masks from the previous stage and the features $\mathbf{F}$ to refine the kernels $\mathbf{K}$ in an iterative way. At every iteration stage $i$, the kernel update iteration operation is formulated as
\begin{equation}
    \mathbf{K}^i, \mathbf{M}^i, \mathbf{z}^i=\mathcal{F}^i(\mathbf{M}^{i-1}, \mathbf{K}^{i-1}, \mathbf{F}),
\end{equation}
    where $\mathbf{z}^i$ is the predicted class label obtained by passing $\mathbf{K}^i$ to a classification head.
    The initial kernels subsequently pass all iteration stages $\mathcal{F}^i(\cdot)$ to learn their parameters during the training period when general image-agnostic localization and shape priors can be encoded. The iteration stage $\mathcal{F}^i(\cdot)$ is identical to the one used in K-Net~\cite{knet}.
    Finally, by the prediction head, we obtain the final prediction $\mathbf{M}^S, \mathbf{z}^S$. The pseudo masks are leveraged to supervise the pre-training of the model. Except for the mask loss and classification loss, an auxiliary loss is responsible for preventing the overfitting of vision features.
    The concrete algorithm of UPLVP is shown in Algorithm~\ref{alg:alg1}.

\subsection{Mask Proposal Generation}
\label{section:mask-generation}
    In this section, we mainly describe how we generate our mask proposals with language-vision models, CLIP and OpenSeg.
    Most previous deep unsupervised segmentation methods generate pseudo masks with unsupervised algorithms for supervised training of the model. Here we also follow the same pipeline and additionally utilize the pseudo masks for generating prompts.
    There exist various unsupervised algorithms for generating pseudo masks, such as selective search~\cite{selective}, random proposal~\cite{updetr}, FreeSOLO~\cite{wang2022freesolo}, and saliency mechanism~\cite{liu2020picanet,liu2021visual,zhuge2022salient,fang2021densely,upsp}.
    While it is simple to obtain pseudo masks using these methods, it is important to note that they only model the dense masks through object separation. This means that the pseudo labels can be annotated by category. Compared to real ground-truth labels, it is obvious that the pseudo labels lack abundant category information. Since the categories can bring rich semantic information that is essential for instance segmentation task, we leverage language-vision models to integrate the category information into our pseudo labels and get high-quality mask proposals. 
    
    In this task, a simple method is used to leverage the pre-trained CLIP~\cite{clip} and OpenSeg~\cite{openseg} models. Our main idea is to generate dense masks by combining text and image features extracted by the two models. Specifically, given an input image $\mathbf{I} \in \mathbb{R}^{H \times W \times 3}$, it is quite easy to obtain its per-pixel features map $\mathbf{X}^I \in \mathbb{R}^{H \times W \times D}$ from an off-the-shelf segmentation model, where $D$ is dimension of extracted features. Here we use OpenSeg as an image encoder $\varepsilon^{I}$. Since we only require OpenSeg to generate a per-pixel features map without class information to ensure it is a class-agnostic segmentation branch, we set text embeddings to zero. 
    
    Each dataset has a certain number of pre-defined categories like COCO has 80 categories. These class names have a significant effect in helping segment pseudo masks since the language-vision model exactly aligns text and image features. Here we leverage pre-trained CLIP as a text encoder $\varepsilon^{T}$ to encode these categories names $\mathbf{T}$ from text to embeddings, denoted as $\mathbf{X}^T \in \mathbb{R}^{D \times C}$, where $D$ represents dimension of feature embeddings and $C$ represents the number of categories of dataset: 
    \begin{equation}
    \mathbf{X}^I=\varepsilon^{I}\left(\mathbf{I}\right) \in \mathbb{R}^{H \times W \times D},
    \vspace{-8pt}
    \end{equation}
    \begin{equation}
    \mathbf{X}^T=\varepsilon^{T}\left(\mathbf{T}\right) \in \mathbb{R}^{D \times C}.
    \end{equation}
    After obtaining text and image features, we linearly normalize two features respectively to get a better alignment. Subsequently, we simply conduct math multiplication of text and image features to predict pseudo masks $\mathbf{\widetilde{M}} \in \mathbb{R}^{H \times W \times C}$, where each channel represents features corresponding to a specific category. Different from some previous work that uses a threshold to filter out unqualified masks with low IoUs, we align the image features with corresponding text embeddings directly instead of randomly generating candidate vision prompts in advance. The high scores of pixels in each channel mean the text-image features pair is highly matched, thus no redundant masks are produced. The whole procedure can be formulated as 
    \begin{equation}
    \mathbf{\widetilde{M}}=\operatorname{Norm}\left(\mathbf{X}^I\right)\cdot\operatorname{Norm}\left(\mathbf{X}^T\right),
    \end{equation}
    where $\operatorname{Norm}(\cdot)$ means linear normalization and $\cdot$ means matrix multiplication. 
    The $c^{th}~(c = \{1,...,C\})$ channel of $\mathbf{\widetilde{M}}$ represents the class label of the mask $\mathbf{\widetilde{M}}_c \in \mathbb{R}^{H \times W}$ and thus we obtain the pseudo labels $\mathbf{\widetilde{z}}$ of all generated masks.
    The process of generating mask proposals is shown in Figure~\ref{fig:overview}.
    
\subsection{Prompt-Kernel Matching}
\label{section:prompt}

    Figure~\ref{fig:overview} shows the details of our proposed Prompt-Kernel Matching, which has two key steps: Prompt Generation and Cosine Similarity Based Matching.

\subsubsection{Prompt generation} 
    Once the pseudo mask proposals are obtained, we can use them to supervise the pre-training directly. However, such a naive strategy would underestimate the potential value of the pseudo masks.
    By revisiting the proposal generation process, we realize that the image features may contain elements or representations similar to their corresponding foreground objects. These features can influence the kernels and encourage them to learn localization and shape priors quickly during the pre-training phase. We propose to transfer the mask proposals, which contain abundant regional knowledge of foreground objects, to visual prompts and inject them into initialized kernels. This will help the kernels to understand object features more quickly, since in our framework the mask predictions are generated by conducting convolution between kernels and feature maps.

    Concretely, given the mask proposals $\mathbf{\widetilde{M}}$ and the image feature maps output by an FPN~\cite{fpn}, we leverage the tightest bounding boxes of proposals to crop the feature maps. The cropped features are denoted as $\mathbf{f}_l = \{\mathbf{f}_1, \mathbf{f}_2, \cdots, \mathbf{f}_L\}\in \mathbb{R}^{h \times w \times D'}$, where $L$ is the number of the masks in $\mathbf{\widetilde{M}}$ ($L$ can vary from different images), $h$ and $w$ are shapes of bounding boxes of pseudo masks and $D'$ is the channel number of the FPN feature. 
    Then, average pooling is operated to convert the features $\mathbf{f}$ into prompts:
    \begin{equation}
    \mathbf{P} = \operatorname{Avg}(\mathbf{f}_l)\in \mathbb{R}^{L \times D'},
    \end{equation}
    where $\operatorname{Avg}(\cdot)$ means average pooling along the spatial dimension.
    
\subsubsection{Cosine similarity based matching}
    As we discussed above, each prompt encodes the localization and shape priors for an individual object and can be injected into one of the initial kernels $\mathbf{K}^0$ of K-Net.
    This raises an interesting question: \emph{which prompt should be injected into which kernel?}
    A straightforward way is to randomly or sequentially assign the $L$ prompts to $N$ tokens. However, DETR~\cite{detr} and K-Net~\cite{knet} have found that under the dynamic learning training scheme, different kernels or queries encode the localization and shape priors of different image regions and objects with varying shapes. Each kernel mainly learns a specific pattern of similar object shapes and locations.
    Therefore, using random or sequential assignments of the prompts may inject completely different object localization and shape information into the same token in different training samples, making the learning of the initial tokens unstable.

    To this end, we propose a novel prompt-kernel matching scheme based on cosine similarity to assign the best-matched prompt to each token.
    Specifically,  given the $L$ prompts $\mathbf{P} = \{\mathbf{P}_l\in \mathbb{R}^{C}\}_{l=1}^L$ and the $N$ initial kernels $\mathbf{K}^0 = \{\mathbf{K}^0_n\in \mathbb{R}^{C}\}_{n=1}^N$, we compute the cosine similarity between them  to build the similarity matrix $\mathbf{E} \in \mathbb{R}^{N\times L}$:
    \begin{equation}
    \mathbf{E}_{n,l} = \frac{\mathbf{K}^0_n}{\|\mathbf{K}^0_n\|_2} \cdot \frac{\mathbf{P}_l}{\left\|\mathbf{P}_l\right\|_2}.
    \end{equation}
    Then, for each kernel $\mathbf{K}^0_n$, we select the best-matched prompt index $\delta(\mathbf{K}^0_n)$ with the largest similarity score:
    \begin{equation}
    \delta(\mathbf{K}^0_n) = \mathop{\arg\max}\limits_{l\in [1,...,L]}\mathbf{E}_{n,l}.
    \end{equation}
    Next, the best-matched prompt $\mathbf{P}_{\delta(\mathbf{K}^0_n)}$ is injected into the kernel $\mathbf{K}^0_n$ via summation:
\begin{equation}
    \mathbf{K}^{0'}_n = \mathbf{K}^0_n + \mathbf{P}_{\delta(\mathbf{K}^0_n)}.
\end{equation}
    Finally, the decorated initial kernels $\mathbf{K}^{0'}$ are fed into the iteration head, where the kernels and masks are iteratively updated. At the $S^{th}$ stage, we obtain the kernels $\mathbf{K}^S$, predicted masks $\mathbf{M}^S$ and predicted class labels $\mathbf{z}^S$.

\begin{table}[t]
\renewcommand{\arraystretch}{1.2}
\caption{
Instance segmentation and fine-tune results on COCO with 5\% and 10\% annotated images based on K-Net. COCO train2017 is used for pre-training of all methods. \vspace{-8pt}
}
\centering
\setlength{\tabcolsep}{1.8mm}{
\begin{tabular}{cccccccc}
\hline
                     & Pre-train                                 & mAP                                   & $AP_{50}$                                & $AP_{75}$                                & $AP_{S}$                                & $AP_{M}$                                 & $AP_{L}$                                 \\ \hline
                             & Img. Sup.~\cite{imagenetdataset}                                   & 14.8                                  & 29.1                                  & 13.7                                  & 4.3                                  & 15.5                                  & 24.4                                  \\
                             & DenseCL~\cite{densecl}                                     & 16.7                                  & 31.2                                  & 15.9                                  & 5.1                                  & 17.5                                  & 27.7                                  \\
                             & SwAV~\cite{swav}                                        & 15.7                                  & 30.3                                  & 14.7                                  & 4.6                                  & 25.9                                  & 16.6                                  \\
                             & MoCo-v2~\cite{mocov2}                                     & 17                                    & 32                                    & 16.2                                  & 5.3                                  & 18.3                                  & 27.1                                  \\
& UPSP~\cite{upsp} & 19.9 & 35.7 & 19.9 & 6.0 & 21.0 & 32.6 \\ 

\multirow{-6}{*}{\rotatebox{90}{5\% imgs}}  & \cellcolor[HTML]{EFEFEF}\textbf{UPLVP} & \cellcolor[HTML]{EFEFEF}\textbf{22.2} & \cellcolor[HTML]{EFEFEF}\textbf{39.5} & \cellcolor[HTML]{EFEFEF}\textbf{22.0} & \cellcolor[HTML]{EFEFEF}\textbf{7.0} & \cellcolor[HTML]{EFEFEF}\textbf{23.4} & \cellcolor[HTML]{EFEFEF}\textbf{37.1} \\ \hline

                             & Img. Sup.~\cite{imagenetdataset}                                   & 19.1                                  & 35.7                                  & 18.2                                  & 6.7                                  & 20                                    & 31.6                                  \\
                             & DenseCL~\cite{densecl}                                     & 20.3                                  & 36.4                                  & 20.3                                  & 6.6                                  & 21.8                                  & 33.6                                  \\
                             & SwAV~\cite{swav}                                        & 18.9                                  & 34.8                                  & 18.3                                  & 6.8                                  & 20.8                                  & 30.6                                  \\
                             & MoCo-v2~\cite{mocov2}                                     & 20.7                                  & 37.7                                  & 20.4                                  & 6.4                                  & 22.1                                  & 34.2                                  \\
& UPSP~\cite{upsp} & 23.5 & 41.4 & 23.7 & 7.9 & 24.8 & 38.6 \\ 

\multirow{-5.5}{*}{\rotatebox{90}{10\% imgs}} & \cellcolor[HTML]{EFEFEF}\textbf{UPLVP} & \cellcolor[HTML]{EFEFEF}\textbf{25.4} & \cellcolor[HTML]{EFEFEF}\textbf{44.4} & \cellcolor[HTML]{EFEFEF}\textbf{25.7} & \cellcolor[HTML]{EFEFEF}\textbf{8.8} & \cellcolor[HTML]{EFEFEF}\textbf{27.2} & \cellcolor[HTML]{EFEFEF}\textbf{42.0} \\ \hline
\end{tabular}}
\vspace{-0.1in}
\label{tab:knet-coco-seg}
\end{table}

\begin{table}[t]
\renewcommand{\arraystretch}{1.2}
\caption{
Object detection fine-tune results on COCO with 5\% and 10\% annotated images based on K-Net. COCO train2017 is used for pre-training of all methods.\vspace{-8pt}
}
\centering
\setlength{\tabcolsep}{3.5mm}{
\begin{tabular}{cccccccc}
\hline
&\multicolumn{4}{c}{Pre-train} & \multicolumn{3}{c}{mAP} \\ \hline
&\multicolumn{4}{c}{DETR $w/$ AlignDet~\cite{aligndet}} & \multicolumn{3}{c}{21.4} \\
&\multicolumn{4}{c}{Faster R-CNN $w/$ AlignDet~\cite{aligndet}} & \multicolumn{3}{c}{22.1} \\
&\multicolumn{4}{c}{Mask R-CNN $w/$ AlignDet~\cite{aligndet}} & \multicolumn{3}{c}{22.4} \\
\multirow{-4}{*}{\rotatebox{90}{5\% imgs}} & \multicolumn{4}{c}{\cellcolor[HTML]{EFEFEF}\textbf{UPLVP}} & \multicolumn{3}{c}{\cellcolor[HTML]{EFEFEF}\textbf{24.2}} \\ \hline

&\multicolumn{4}{c}{DETR $w/$ AlignDet~\cite{aligndet}} & \multicolumn{3}{c}{26.1} \\
&\multicolumn{4}{c}{Faster R-CNN $w/$ AlignDet~\cite{aligndet}} & \multicolumn{3}{c}{26.9} \\
&\multicolumn{4}{c}{Mask R-CNN $w/$ AlignDet~\cite{aligndet}} & \multicolumn{3}{c}{27.4} \\
\multirow{-3.5}{*}{\rotatebox{90}{10\% imgs}} & \multicolumn{4}{c}{\cellcolor[HTML]{EFEFEF}\textbf{UPLVP}} & \multicolumn{3}{c}{\cellcolor[HTML]{EFEFEF}\textbf{27.8}} \\ \hline

\end{tabular}}
\vspace{-0.1in}
\label{tab:knet-coco-det}
\end{table}

\begin{table}[t]
\renewcommand{\arraystretch}{1.2}
\caption{
Instance segmentation fine-tune results on COCO with 5\% and 10\% annotated images based on K-Net. UPLVP is pre-trained with COCO train2017 and FreeSOLO was pre-trained with COCO train2017 and COCO unlabeled2017. \vspace{-8pt}
}
\centering
\setlength{\tabcolsep}{1.8mm}{
\begin{tabular}{cccccccc}
\hline
\multicolumn{8}{c}{Instance Segmentation}\\
\hline
                     & Pre-train                                 & mAP                                   & $AP_{50}$                                & $AP_{75}$                                & $AP_{S}$                                & $AP_{M}$                                 & $AP_{L}$                                 \\ \hline

  \multirow{-0.5}{*}{\rotatebox{90}{5\%}}   & FreeSOLO~\cite{wang2022freesolo}    & 22.0 & 36.0 & \textbf{22.9} & 6.5 & 23.2 & 33.8 \\
  & \cellcolor[HTML]{EFEFEF}\textbf{UPLVP} & \cellcolor[HTML]{EFEFEF}\textbf{22.2} & \cellcolor[HTML]{EFEFEF}\textbf{39.5} & \cellcolor[HTML]{EFEFEF}22.0 & \cellcolor[HTML]{EFEFEF}\textbf{7.0} & \cellcolor[HTML]{EFEFEF}\textbf{23.4} & \cellcolor[HTML]{EFEFEF}\textbf{37.1} \\ \hline

 \multirow{-0.7}{*}{\rotatebox{90}{10\%}} & FreeSOLO~\cite{wang2022freesolo}    & \textbf{25.6} & 41.6 & \textbf{26.7} & 8.3 & \textbf{27.5} & 40.3 \\
  & \cellcolor[HTML]{EFEFEF}\textbf{UPLVP} & \cellcolor[HTML]{EFEFEF}25.4 & \cellcolor[HTML]{EFEFEF}\textbf{44.4} & \cellcolor[HTML]{EFEFEF}25.7 & \cellcolor[HTML]{EFEFEF}\textbf{8.8} & \cellcolor[HTML]{EFEFEF}27.2 & \cellcolor[HTML]{EFEFEF}\textbf{42.0} \\ \hline

\end{tabular}}
\label{UPLVP-freesolo}
\end{table}

\subsection{Loss Function and Kernel Supervision}
\label{section:loss}
    During the pre-training phase, we leverage the generated pseudo masks $\mathbf{\widetilde{M}}$ and $\mathbf{\widetilde{z}}$ as the segmentation ground truth to supervise the pre-training.
    Note that in the fine-tuning phase, the real ground truth is used to supervise the models.
    
    Since the pseudo class labels are generated by conducting math multiplication between vision and text features, we argue that a similar method to predict classes in the vanilla QEIS models is needed to maintain task consistency. 
    It should be noted that the feature space pre-trained for predicting pseudo labels may differ from the one trained to predict COCO ground-truth labels. Therefore, to prevent overfitting of the K-Net part during the pre-training phase, we introduce a supplementary auxiliary loss that supervises the kernels to predict classes. It's important to note that this loss function is only used during pre-training and will not be integrated into the real training phase.
 
    Concretely, updated kernels after the $S^{th}$ kernel iteration head $\mathbf{K}^S = \{\mathbf{K}^S_1, \mathbf{K}^S_2, \cdots, \mathbf{K}^S_N\}$ can be regarded as refined features containing information of corresponding instances. Each kernel is responsible for predicting a mask corresponding to its location and shape.
    The kernels resemble vision features obtained by mask proposal network $\mathbf{X}^I$. At this point, we similarly conduct math multiplication between kernels and text embedding, and the supervision loss is the same as the focal loss. The auxiliary loss can be formulated as follows:
    \begin{equation}
    \mathbf{Q} = \mathbf{K}^S\cdot\mathbf{X}^T \in \mathbb{R}^{N\times C},
    \vspace{-4pt}
    \end{equation}
    \begin{equation}
    \mathbf{a} = \mathop{\arg\max}\limits_{c\in [1,...,C]}\mathbf{Q} \in \mathbb{R}^{N\times 1},
    \vspace{-1pt}
    \end{equation}
    \begin{equation}
    \mathcal{L}_{aux}=-\sum_{\xi}y_j*log(\mathbf{a}_\xi),
    \end{equation}
    where $y_j~(j = \{1,...,N\})$ is pseudo category labels of $\mathbf{\widetilde{M}}_j$, $\mathbf{a}$ is the indices of the categories predicted by all kernels and $\xi$ is the index of the predicted category selected by Hungarian matching algorithm.
    Our final loss  can be defined as
    \begin{equation}
    \begin{split}
    \mathcal{L} = \lambda_{cls}\mathcal{L}_{c l s}(\mathbf{\widetilde{z}}, \mathbf{z}^S) + \lambda_{dice}\mathcal{L}_{dice}(\mathbf{\widetilde{M}}, \mathbf{M}^S) + \\
    \lambda_{ce}\mathcal{L}_{ce}(\mathbf{\widetilde{M}}, \mathbf{M}^S) +\lambda_{aux} \mathcal{L}_{aux}(\mathbf{\widetilde{z}}, \mathbf{a}),
    \end{split}
    \end{equation}
    where $\lambda_{(\cdot)}$ are corresponding loss weights, $\mathcal{L}_{c l s}$ is Focal loss for classification, and $\mathcal{L}_{ce}$ and $\mathcal{L}_{dice}$ are CrossEntropy loss and Dice loss for segmentation. The auxiliary loss process is shown in Figure~\ref{fig:overview}.

\section{Experiments}
\label{experiments}
\subsection{Implementation Details}

\begin{figure}[t]
  \centering
  \includegraphics[width=1\linewidth]{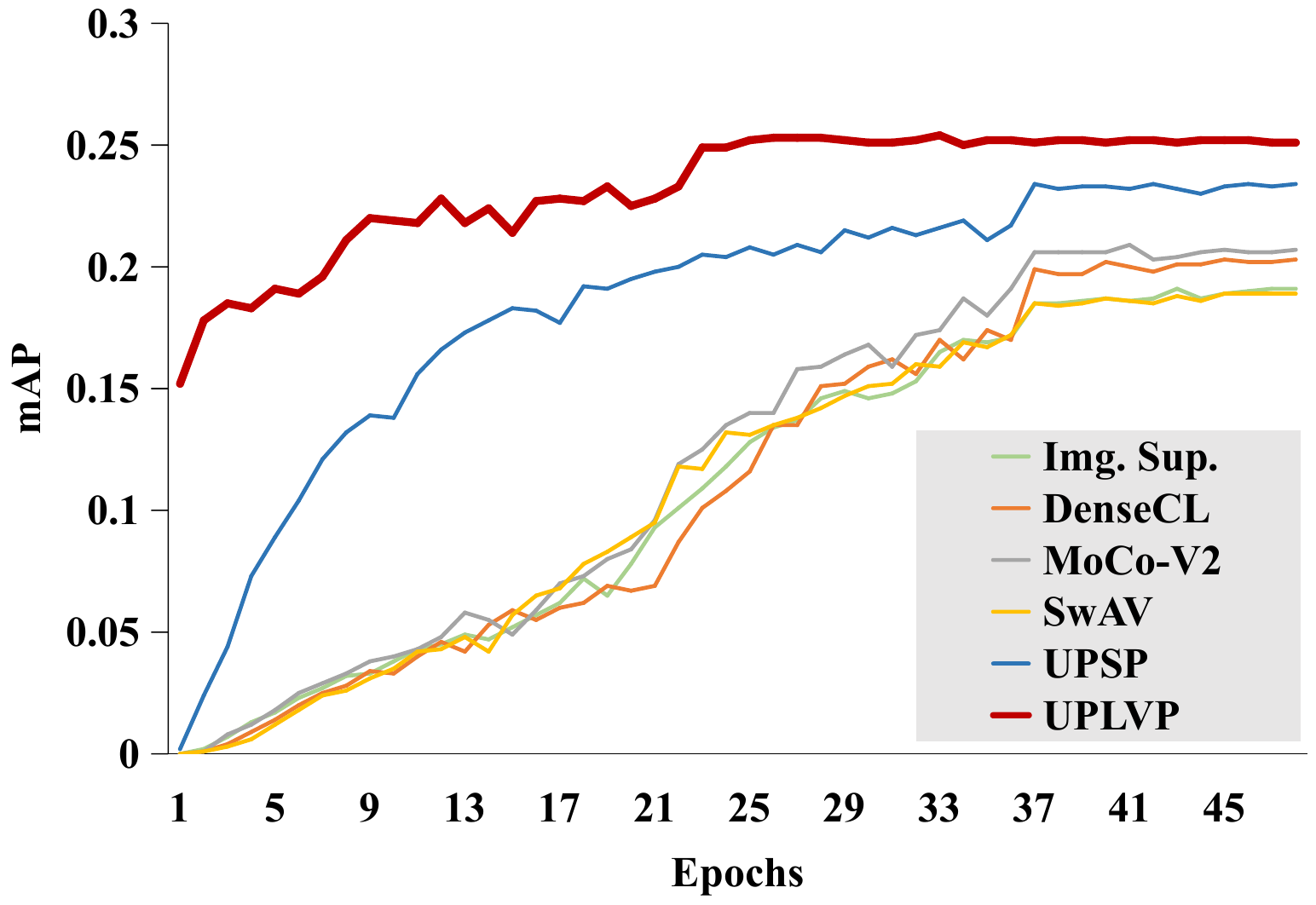}
  \caption{AP learning curves of K-Net with different pre-training methods on COCO with 10\% annotated images. As can be seen, our method can speed up the convergence during fin-tuning phase.}
  \label{fig:10image}
  \vspace{-10pt}
\end{figure}

\label{sec:exp}
\subsubsection{Pre-training setting}
    We adopt ResNet-50~\cite{resnet} for all models as the backbone and pre-trained with the DenseCL algorithm.
    The AdamW optimizer is adopted with $0.05$ weight decay and 1,000 steps of linear step warmup. 
    As for data augmentation, we simply apply random flipping.
    The model is pre-trained with a batch size of 96 for 12 epochs on 8 A100 GPUs.
    The initial learning rate is set to $1\times 10^{-4}$ and decreased by 0.1 after 8 and 11 epochs.
    As for the hyper-parameters of our model, we set the number of kernels/queries $N$ as 100,  $\mathcal{L}_{c l s}=2$,  $\mathcal{L}_{dice}=4$, $\mathcal{L}_{ce}=1$ and $\mathcal{L}_{aux}=2$.

\subsubsection{Fine-tuning setting} 
    All models are trained with a batch size of 96 on 8 A100 GPUs.
    Random flipping and rotation are used as data augmentation.
    Referring to the open-source of MMDetection~\cite{mmdetection}, the same hyper-parameters are used for QEIS models and CNN-based models.
    For QEIS models, we use the same training strategy as the pre-training stage except for the training epoch, which will be illustrated in the following experiment tables. 
    For CNN-based models, we apply SGD as the optimizer with weight decay and momentum. The learning rate is 0.02, momentum is set to 0.9 and weight decay is 0.0001.

\begin{table*}[t]
\renewcommand{\arraystretch}{1.2}
\centering
\caption{Instance segmentation fine-tune results on Cityscapes and CTW1500. COCO train2017 is used for pre-training. As can be seen, our pre-training method can improve the ability of QEIS models to segment novel instances.}\vspace{-8pt}
\label{tab:knet-other}
\setlength{\tabcolsep}{1.9mm}{
\begin{tabular}{cc|cccccc|ccccccc}
\hline
                        &                                           & \multicolumn{6}{c|}{Cityscapes}                                                                                                                                                                                                                                    & \multicolumn{7}{c}{CTW1500}                                                                                                                                                                                                                                            \\
\multirow{-2}{*}{Model} & \multirow{-2}{*}{Pre-train}               & Epoch                               & AP                                    & $AP_{50}$                                                    & $AP_{S}$                    & $AP_{M}$                            & $AP_{L}$                            & Epoch                               & AP                                    & $AP_{50}$                             & $AP_{75}$                           & $AP_{S}$                   & $AP_{M}$                              & $AP_{L}$                              \\ \hline
Mask RCNN~\cite{maskrcnn}               & Img. Sup.~\cite{imagenetdataset}                                      & 24                                  & 30                                    & \textbf{57.4}                                                                    & \textbf{8.3}                         & 27.9                                & 49                                  & 96                                  & 34.5                                  & 69.8                                  & 32.1                                & \textbf{25.9}                       & 40.4                                  & 36.8                                  \\
SOLO-v2~\cite{solov2}                 & Img. Sup.~\cite{imagenetdataset}                                      & 24                                  & 24.9                                  & 44.4                                                                    & 1.8                         & 20.1                                & 50.4                                & 96                                  & 27.9                                  & 59.6                                  & 23.3                                & 8.7                        & 30.2                                  & 41                                    \\ \hline
                        & Img. Sup.~\cite{imagenetdataset}                                      & 24                                  & 24.8                                  & 47.4                                                                    & 4.8                         & 19.9                                & 43.1                                & 96                                  & 9.7                                   & 26.5                                  & 6.1                                 & 3.0                          & 9.2                                   & 19.2                                  \\
                        & DenseCL~\cite{densecl}                                   & 24                                  & 28                                    & 52.2                                                                    & 6.6                         & 25.2                                & 55.2                                & 96                                  & 18.9                                  & 42.6                                  & 15.1                                & 7.0                          & 18.8                                  & 32.5                                  \\
                        & SwAV~\cite{swav}                                      & 24                                  & 27.4                                  & 52.1                                                                    & 5.3                         & 22.7                                & 49.2                                & 96                                  & 9.1                                   & 25.8                                  & 4.8                                 & 2.7                        & 9                                     & 19.8                                  \\
                        & MoCo-v2~\cite{mocov2}                                   & 24                                  & 28.2                                  & 51.2                                                                    & 5.9                         & 26.4                                & 52.7                                & 96                                  & 13.3                                  & 32.2                                  & 10                                  & 4.3                        & 13.3                                  & 24.2                                  \\
 & UPSP~\cite{upsp} & 24 & 30.6 & 55.4 & 5.8 & 27.0 & 54.0 & 96 & 34.6 & 71.1 & 31.0 & 18.0 & 36.1 & 45.7 \\ 
 
\multirow{-6}{*}{K-Net~\cite{knet}} & \cellcolor[HTML]{EFEFEF}\textbf{UPLVP} & \cellcolor[HTML]{EFEFEF}\textbf{24} & \cellcolor[HTML]{EFEFEF}\textbf{32.4} & \cellcolor[HTML]{EFEFEF}57.3 & \cellcolor[HTML]{EFEFEF}7.3 & \cellcolor[HTML]{EFEFEF}\textbf{28.8} & \cellcolor[HTML]{EFEFEF}\textbf{57.8} & \cellcolor[HTML]{EFEFEF}\textbf{96} & \cellcolor[HTML]{EFEFEF}\textbf{39.9} & \cellcolor[HTML]{EFEFEF}\textbf{74.8} & \cellcolor[HTML]{EFEFEF}\textbf{40.3} & \cellcolor[HTML]{EFEFEF}23.9 & \cellcolor[HTML]{EFEFEF}\textbf{42.3} & \cellcolor[HTML]{EFEFEF}\textbf{50.6} \\ \hline
\end{tabular}}
\vspace{-10pt}
\end{table*}

\begin{figure}[t]
    \centering
    \subfloat[AP learning curves of Cityscapes.]{
        \includegraphics[width=1\linewidth]{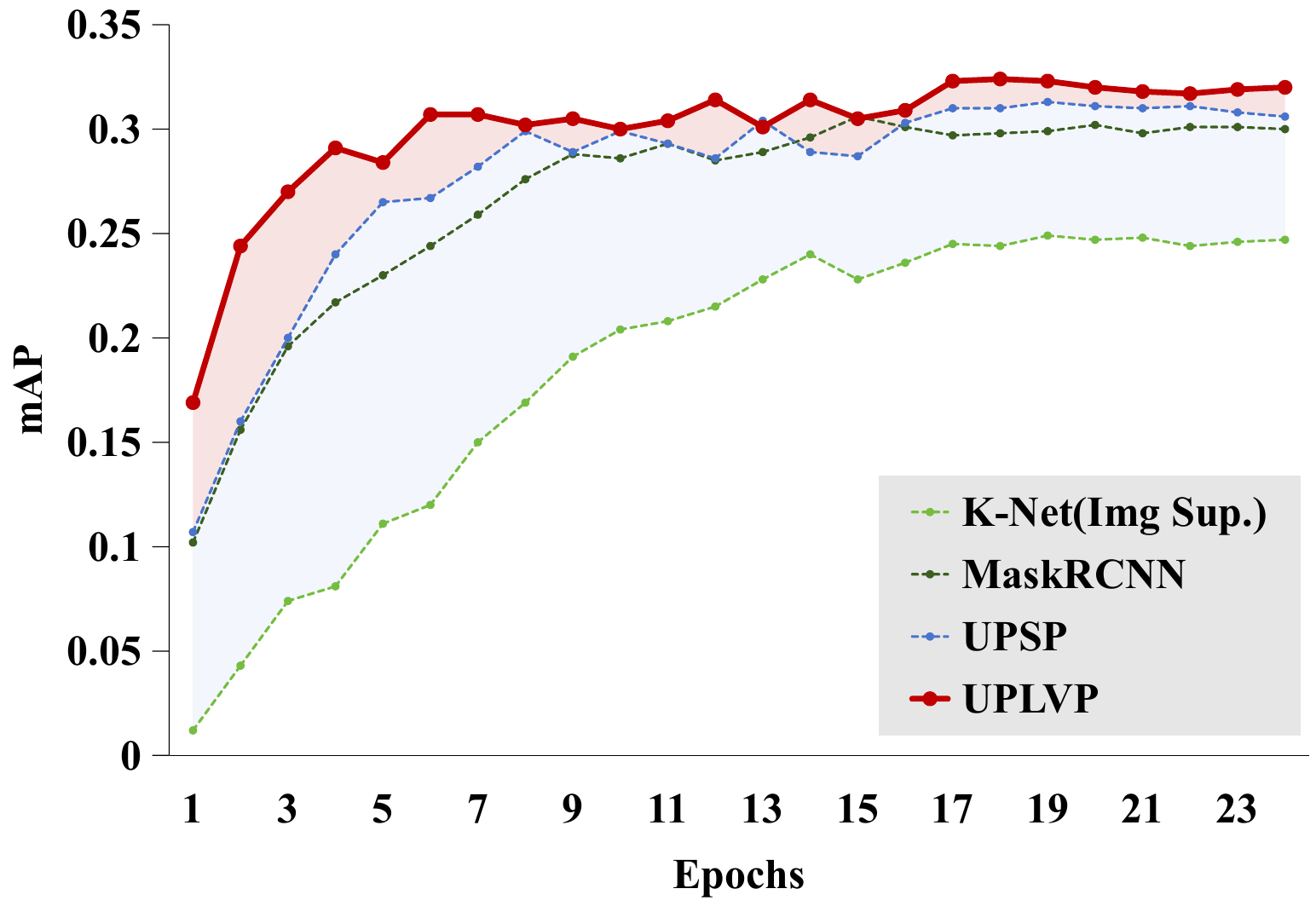}
    }

    \subfloat[AP learning curves of CTW1500]{
        \includegraphics[width=1\linewidth]{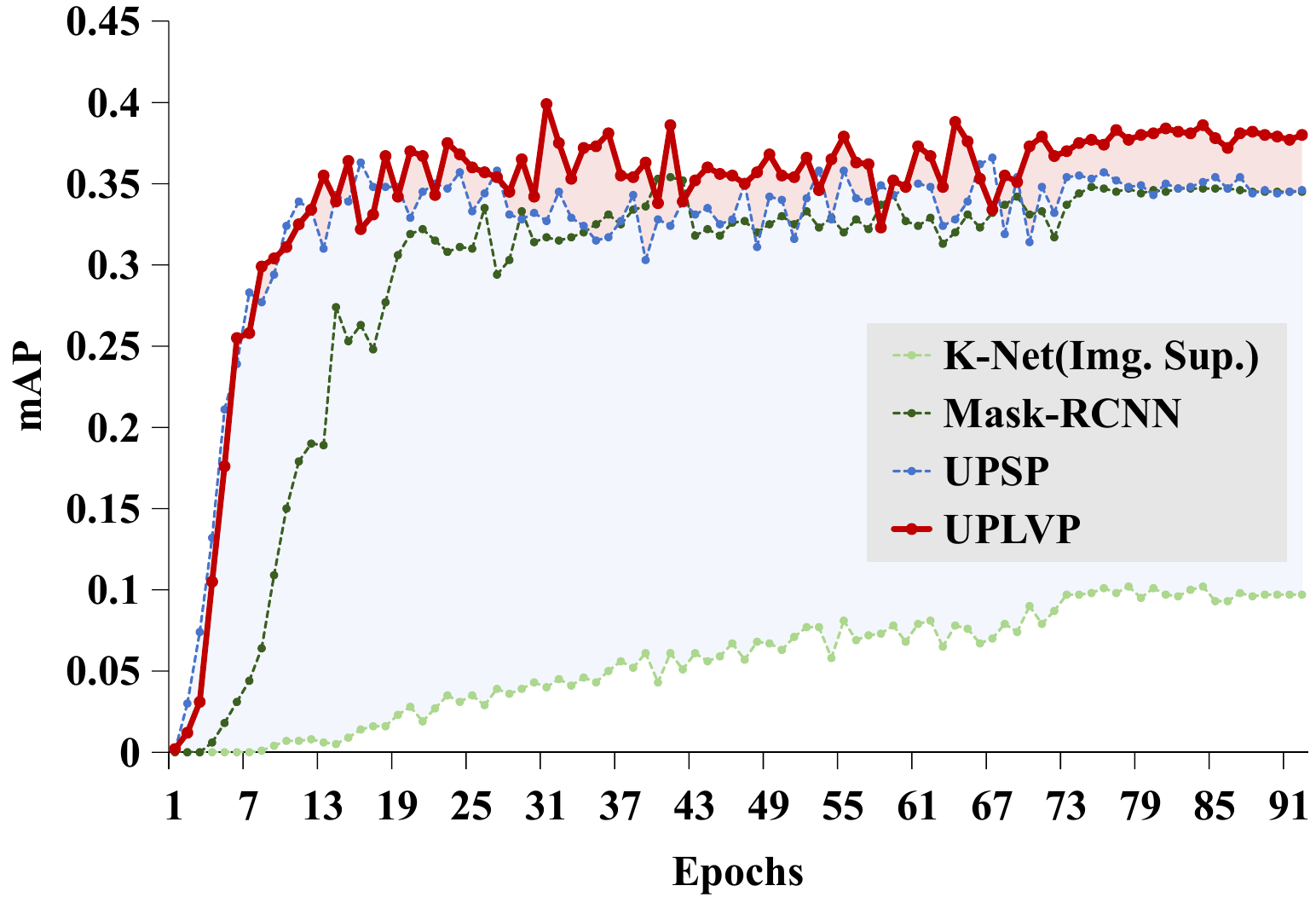}
    }
    \caption{AP learning curves of Mask-RCNN, vanilla K-Net and K-Net pre-trained with different methods. COCO train2017 is used for pre-training, Cityscapes and CTW1500 are used for fine-tuning for all methods.}
    \label{fig:curve-knet-other}
    \vspace{-10pt}
\end{figure}

\subsubsection{Datasets} 
    Our approach involves two main phases: the pre-training phase and the fine-tuning phase. In the pre-training phase, we train the model with a specific dataset, and in the fine-tuning phase, we refine the model's performance with the same or another dataset. To evaluate the effectiveness of our pre-trained QEIS models, we conduct experiments in two parts.: \emph{1) the same dataset is used in the pre-training and fine-tuning phase to test with seen objects; 2) different datasets are used in the pre-training and fine-tuning phase to test with unseen objects.}
    In general, our QEIS models are pre-trained on MS COCO~\cite{lin2014coco} unlabeled 2017 split and fine-tuned on multiple datasets, including MS COCO, Cityscapes~\cite{cityscapes} and CTW1500~\cite{ctw1500}. In task of few-shot instance segmentation, we further test our model in PASCAL VOC2012.
    {MS COCO} is a popular instance segmentation dataset that contains 164k labeled images, where objects of 80 object categories are annotated with dense pixel segmentation.
    {Cityscapes} is a popular instance segmentation dataset that focuses on semantic understanding and urban street scenes. It contains 5000 fine annotated large-scale images.
    {CTW1500} is a wild-scene dataset that focuses on text detection and segmentation. It contains 1,500 images with dense annotation, which is also a typical low-data regime.
    {VOC2012 Object Segmentation} includes 20 classes that are a subset of COCO's 80 classes. The training set consists of 1464 images and the validation set has 1449 images.

\subsubsection{Evaluation metrics}
    In order to assess the effectiveness of our UPLVP models, we use the same evaluation metrics that are used in previous studies on instance segmentation, which involves measuring the Average Precision (AP) of the results. We measure the mean Average Precision (mAP) of each experiment to evaluate the overall ability of instance segmentation and consider mAP as the main metric. For small, medium, and large objects, we report their AP as $AP_s$, $AP_m$, and $AP_l$, respectively. The AP is reported as $AP_{50}$ at a mask IoU threshold of 0.5 and as $AP_{75}$ at a mask IoU threshold of 0.75. It is worth noting that when using Cityscapes as a fine-tune dataset, we do not evaluate $AP_{75}$.

\begin{table*}[t]
\renewcommand{\arraystretch}{1.2}
\caption{
Instance segmentation fine-tune results using QueryInst and Mask2Former on Cityscapes and CTW1500. COCO train2017 is used for pre-training for all methods. As can be seen, our pre-training method is feasible to be applied to QEIS models.
}\vspace{-8pt}
\centering
\setlength{\tabcolsep}{1.9mm}{
\begin{tabular}{cc|ccccccc|cccccc}
\hline
                              &                                             & \multicolumn{7}{c|}{CTW1500}                                                                                                                                                                                                                                                         & \multicolumn{6}{c}{Cityscapes}                                                                                                                                                                                                                                                 \\
\multirow{-2}{*}{Model}       & \multirow{-2}{*}{Pre-train}                 & Epoch                               & AP                                    & $AP_{50}$                             & $AP_{75}$                             & $AP_{S}$                              & $AP_{M}$                              & $AP_{L}$                              & Epoch                               & AP                                    & $AP_{50}$                                                    & $AP_{S}$                             & $AP_{M}$                              & $AP_{L}$                             \\ \hline
                              & Img. Sup.~\cite{imagenetdataset}                                        & 80                                  & 38.8                                  & 67.6                                  & 41.6                                  & 15.6                                  & 41.2                                  & 57.4                                  & 24                                  & 29.1                                  & 52.4                                                                    & 5.6                                  & 23.9                                  & 55                                   \\
                              & DenseCL~\cite{densecl}                                      & 80                                  & 43.2                                  & 71.6                                  & 48.5                                  & 18.4                                  & 47.6                                  & 59.9                                  & 24                                  & 27.5                                  & 48.9                                                                    & 5.2                                  & 23.8                                  & 53.7                                 \\
                              & SwAV~\cite{swav}                                         & 80                                  &   41.2                            &      69.1                       &       46.1                       &      17.6                          &       45.2                          &        58.1                        & 24                                  & 30.3                                  & 53.3                                                                    & 5.4                                  & 23.3                                  & 59                                   \\
                              & MoCo-v2~\cite{mocov2}                                      & 80                                  & 43.3                                  & 71.2                                  & 49.2                                  & 18.9                                  & 47.6                                  & 59.4                                  & 24                                  & 30.7                                  & 54.3                                                                    & 5.4                                  & 25.5                                  & 56.4                                 \\
 & UPSP~\cite{upsp}  & 20 & 52.9 & 83.4 & 62.1 & 29.4 & 56.4 & 67.6 & 24 & 31.8 & 55.8 & 5.1 & 26.5 & 59.0 \\ 
 
\multirow{-6}{*}{Mask2Former~\cite{mask2former}} & \cellcolor[HTML]{EFEFEF}\textbf{UPLVP} & \cellcolor[HTML]{EFEFEF}\textbf{20} & \cellcolor[HTML]{EFEFEF}\textbf{54.1} & \cellcolor[HTML]{EFEFEF}\textbf{84.3} & \cellcolor[HTML]{EFEFEF}\textbf{63.3} & \cellcolor[HTML]{EFEFEF}\textbf{31.2} & \cellcolor[HTML]{EFEFEF}\textbf{56.9} & \cellcolor[HTML]{EFEFEF}\textbf{70.3} & \cellcolor[HTML]{EFEFEF}\textbf{24} & \cellcolor[HTML]{EFEFEF}\textbf{34.2} & \cellcolor[HTML]{EFEFEF}\textbf{58.2}  & \cellcolor[HTML]{EFEFEF}\textbf{7.1} & \cellcolor[HTML]{EFEFEF}\textbf{28.8} & \cellcolor[HTML]{EFEFEF}\textbf{62.4} \\ \hline
                              & Img. Sup.~\cite{imagenetdataset}                                         & 80                                  & 28.3                                  & 53.7                                  & 28.6                                  & 9.8                                   & 29                                    & 41.8                                  & 24                                  & 29.1                                  & 53.2                                                                    & 6.7                                  & 27.4                                  & 50.7                                 \\
                              & DenseCL~\cite{densecl}                                      & 80                                  & 31.6                                  & 56.7                                  & 33.4                                  & 10.4                                  & 32.5                                  & 46.6                                  & 24                                  & 30.8                                  & 54.7                                                                    & 8.6                                  & 28.9                                  & 54.5                                 \\
                              & SwAV~\cite{swav}                                         & 80                                  & 24.6                                  & 50                                    & 23.1                                  & 8.1                                   & 25                                    & 36.3                                  & 24                                  & 30.7                                  & 54.4                                                                    & 7.9                                  & 28.5                                  & 53.9                                 \\
                              & MoCo-v2~\cite{mocov2}                                      & 80                                  & 31.6                                  & 56.8                                  & 32.8                                  & 12.6                                  & 32                                    & 45.8                                  & 24                                  & 31.4                                  & 54.4                                                                    & 8.1                                  & 28.4                                  & 56.1                                 \\
 & UPSP~\cite{upsp}  & 20 & 39.2 & 66.8 & 43.1 & 16.7 & 42.2 & 51.9 & 24 & 32.8 & 57.3 & 8.8 & 29.2 & 57.0  \\   
\multirow{-6}{*}{QueryInst~\cite{queryinst}}  & \cellcolor[HTML]{EFEFEF}\textbf{UPLVP} & \cellcolor[HTML]{EFEFEF}\textbf{20} & \cellcolor[HTML]{EFEFEF}\textbf{43.2} & \cellcolor[HTML]{EFEFEF}\textbf{71.2} & \cellcolor[HTML]{EFEFEF}\textbf{48.1} & \cellcolor[HTML]{EFEFEF}\textbf{23.6} & \cellcolor[HTML]{EFEFEF}\textbf{46.4} & \cellcolor[HTML]{EFEFEF}\textbf{57.8} & \cellcolor[HTML]{EFEFEF}\textbf{24} & \cellcolor[HTML]{EFEFEF}\textbf{35.2} & \cellcolor[HTML]{EFEFEF}\textbf{59.9}  & \cellcolor[HTML]{EFEFEF}\textbf{8.7} & \cellcolor[HTML]{EFEFEF}\textbf{31.9} & \cellcolor[HTML]{EFEFEF}\textbf{61.4}  \\ \hline
\end{tabular}}
\vspace{-10pt}
\label{tab:query-mask2former}
\end{table*}

\subsection{Fine-tune Results on MS COCO}
    To evaluate our performance in low-data regimes, we split the MS COCO train2017 dataset into two different types of training subset: 1) COCO with 10\% fully annotated images containing 12k+ images, 80k+ annotated masks; 2) COCO with 5\% fully annotated images containing 5k+ images, 43k+ annotated masks.
    Table~\ref{tab:knet-coco-seg} presents the comparison results of various pre-training methods on MS COCO. The term Img. Sup. represents ImageNet supervised pre-training. UPSP~\cite{upsp} obtains pseudo masks and transfers them into prompts to help kernels learn localization and shape priors. However, it only adopts a saliency mechanism to generate mask proposals, which includes foreground and background labels while excluding category information.
    
    The performance of vanilla K-Net is poor in low-data scenarios even when using pre-training methods like Img. Sup., DenseCL, SwAV, and MoCo-v2. These pre-training weights only pre-train the backbones and neglect the instance segmentation head, which means the weights of kernels and kernel update iteration head in K-Net remain initialized to all zeros or random numbers. These are critical parts of K-Net in instance segmentation. 
    Our pre-training method, however, significantly improves performance compared to ImageNet supervised pre-training: an increase by \textbf{+7.4} mAP on 5\% COCO and \textbf{+6.3} mAP on 10\% COCO. The proposed UPLVP, where kernels and kernel update iteration heads in K-Net are mainly pre-trained, is effective and outperforms other pre-training methods where backbones are mainly pre-trained.
    We compare our pre-training method to the UPSP, which pre-trains kernels and segmentation heads. Our method shows an additional increase in mAP of +2.3 on 5\% COCO and +1.9 on 10\% COCO, demonstrating the effectiveness of integrating category knowledge. 
    Subsequently, we convert our instance segmentation masks into bounding boxes and evaluated the results using AlignDet~\cite{aligndet}, a method for object detection. The comparison results with Mask R-CNN using AlignDet are presented in table~\ref{tab:knet-coco-det}. Our method achieved a higher mAP score of additional \textbf{+2.0} on 5\% COCO and additional \textbf{+0.4} on 10\% COCO.

    In our study, we compare the UPLVP and FreeSOLO pre-training techniques and have presented the results in table~\ref{UPLVP-freesolo}. For pre-training, we use COCO train2017 and COCO unlabeled2017 datasets for FreeSOLO whereas for UPLVP, only COCO train2017 is used. Both methods are fine-tuned with COCO 5\% and 10\% train2017 images. Despite using fewer images for pre-training as compared to FreeSOLO, our method has achieved AP results that are equivalent to those of FreeSOLO.

    Moreover, Figure~\ref{fig:10image} shows the learning curves of AP on 10\% COCO images. From these curves, we can conclude that UPLVP converges much faster than the backbones pre-trained methods, and at the beginning of the fine-tuning stage, it gains much higher AP and outperforms UPSP. UPLVP has already reached 0.15 mAP at the first epoch, which is about 10 epochs faster than UPSP and 30 epochs faster than other methods.
    These results suggest that our method has learned shape and localization priors in the pre-training stage, and it provides useful information to the instance segmentation head.

\subsection{Fine-tune Results on Other Datasets}
    In this section, we continue to pre-train our model using the MS COCO dataset and evaluate its performance on wild scenes with new targets (CTW1500~\cite{ctw1500}) and small objects (Cityscapes~\cite{cityscapes}). The comparison results are presented in Table~\ref{tab:knet-other}. 
    Our UPLVP method outperforms the ImageNet-supervised approach by \textbf{+7.6} AP, and achieves better performance (\textbf{+2.4} AP) compared to Mask-RCNN, a popular CNN-based model, on Cityscapes. Additionally, our method outperforms the UPSP approach by \textbf{+1.8} AP. On CTW1500, our method outperforms both supervised and unsupervised methods by \textbf{+30.2} AP and \textbf{+21.0} AP, respectively. It also outperforms the UPSP approach by \textbf{+5.3} AP, and achieves better performance (\textbf{+5.4} AP) with Mask-RCNN.

    These experiments demonstrate that our method can help QEIS models learn localization and shape priors, rather than just remembering objects during the pre-train phase. However, in scenarios involving small objects, our method did not show superior results when compared to Mask-RCNN in terms of $AP_{S}$. We believe there are two reasons for this. Firstly, the QEIS paradigm may have intrinsic deficiencies, as both vanilla SOLOv2 and K-Net perform extremely poorly compared with Mask R-CNN. Secondly, Mask Proposals mainly provide large-scale pseudo labels, which makes our kernels pay more attention to big objects rather than small objects. Although our proposed UPLVP cannot achieve higher $AP_{S}$ results than Mask R-CNN, it is still better than UPSP since more small-scale pseudo mask proposals are generated with the help of language-vision models. From this perspective, UPLVP can compensate for the deficiency of UPSP.

    The AP learning curves on Cityscapes and CTW1500 are shown in Figure~\ref{fig:curve-knet-other}. 
    Although QEIS models like K-Net perform much worse than traditional CNN models like Mask-RCNN in both convergency speed and final accuracy,
    equipped with our method, K-Net converges much faster than the ImageNet supervised pre-training models and can gain considerable learning curves compared with Mask-RCNN. 

\subsection{Deployment on QueryInst and Mask2Former}
    In this section, we apply our pre-training method to two additional QEIS models, QueryInst~\cite{queryinst} and Mask2Former~\cite{mask2former}, to further validate its compatibility with QEIS models.
    For Mask2former and QueryInst, we similarly sample regional features with bounding boxes of pseudo masks and transfer them into language-vision prompts. These prompts are then added to their best-matched queries to decode localization and shape of instances.
    As shown in table~\ref{tab:query-mask2former}, using our pre-training method UPLVP, QueryInst outperforms the current state-of-the-art backbone pre-training method by \textbf{+11.6} AP on CTW1500 dataset and achieves gains of \textbf{+3.8} AP on Cityscapes dataset. Similarly, compared to the UPSP~\cite{upsp}, there are gains of \textbf{+4.0} AP on CTW1500 dataset and \textbf{+2.4} AP on Cityscapes dataset. 
    For Mask2Former, our method achieves significant gains of \textbf{+10.8} AP over the state-of-the-art backbone pre-training method on CTW1500 and gains of \textbf{+3.5} AP on Cityscapes dataset. Compared with UPSP, there are gains of \textbf{+1.2} AP on CTW1500 dataset and \textbf{+2.4} AP on Cityscapes dataset. 
    These results indicate that our pre-training method is highly compatible with QEIS models, which allows the kernels/queries of QEIS models to learn localization and shape priors effectively, leading to significant performance improvements.

\begin{figure}[t]
  \centering
  \includegraphics[width=1\linewidth]{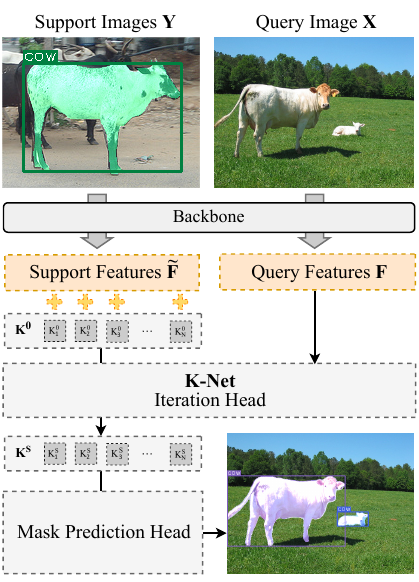}
  \caption{Pipeline of the few-shot instance segmentation task. The query image and the support images are fed into the shared backbone and the extracted support features are injected into all kernels. In this experiment, we adopt COCO2VOC setting where we train models on the subset of COCO train2017 and test them on VOC2012.}
  \label{fig:fewshot}
  \vspace{-10pt}
\end{figure}



\begin{table}[t]
\renewcommand{\arraystretch}{1.2}
\caption{
Performance in terms of $AP_{50}$ obtained by different methods under the COCO2VOC setting. Compared with vanilla K-Net, our pre-training method can improve the performance of QEIS models in the few-shot task.
}\vspace{-8pt}
\centering
\begin{tabular}{ccc}
\hline
   \multicolumn{3}{c}{Instance Segmentation}   \\ \hline
Method & 1way-1shot    & 3way-3shot  \\  \hline
MRCNN-FT~\cite{fgn}  & 0.4    & 2.7  \\
Siamese MRCNN~\cite{siamese}  & 13.8    & 6.6  \\
Meta R-CNN~\cite{metarcnn}  & 12.5   & 15.3  \\
FGN~\cite{fgn}  & 16.2    & 17.9  \\ 
K-Net~\cite{knet}  & 15.8    & 17.1  \\ 
\cellcolor[HTML]{EFEFEF}\textbf{K-Net $w/$ UPLVP}   & \cellcolor[HTML]{EFEFEF}\textbf{18.3}    & \cellcolor[HTML]{EFEFEF}\textbf{18.8}  \\ \hline

\end{tabular}
\vspace{-10pt}
\label{tab:fewshot}
\end{table}

\begin{table}[t]
\renewcommand{\arraystretch}{1.2}
\caption{Ablation of {kernel supervised learning}. '\XSolidBrush' means we remove the auxiliary loss during the pre-training phase and the loss function is the same with vanilla K-Net.} \vspace{-8pt}
\centering
\begin{tabular}{cccccccc}
\hline
Model          & $L_{aux}?$           & mAP           & $AP_{50}$        & $AP_{75}$        & $AP_{S}$        & $AP_{M}$         & $AP_{L}$         \\ \hline
\rowcolor[HTML]{EFEFEF} 
\textbf{UPLVP} & {\color[HTML]{009901}\textbf{\Checkmark}} & \textbf{25.4} & \textbf{44.4} & \textbf{25.7} & 8.6 & \textbf{27.2} & \textbf{42.0} \\
UPLVP          & {\color[HTML]{CB0000}\XSolidBrush}        & 25.3          & 44.3          & 25.5          & \textbf{8.8}          & 27.1          & 41.6          \\ \hline
\end{tabular}
\label{tab:loss_sem-ablation}
\vspace{-10pt}
\end{table}

\subsection{Fine-tune for Few-Shot Instance Segmentation}
    Given an episode $\mathcal{E} = \{(\mathbf{X}, \mathbf{Y}_i)|~i = 1,...,N \times K\}$ consists of an input query image $\mathbf{X}$ and a few support images $\mathbf{Y}$, the task of few-shot instance segmentation is to segment all instances in the query image belonging to the novel classes of the support images. The number of support images formulates the $N-way~K-shot$ problem, where $N$ represents the number of novel classes and $K$ represents the number of annotated instances for each class.

    During the fine-tuning phase, we randomly select a query image and $N$ categories that appear in the query image. Then $K$ images for each category are randomly selected, resulting in $N \times K$ support images. 
    We then extract the feature map $\mathbf{\widetilde{F}}_n^k~(n \in \{1,...,N\}, k \in \{1,...,K\})$ for each support image using the CNN backbone shared with the query image. For each class $n$, we average the features $\mathbf{\widetilde{F}}_n$ of $K$ support images and add them to all kernels.
    To evaluate how well our models can generalize to new data, we adopt a cross-dataset approach~\cite{fgn}, which involves using data from both MS COCO and PASCAL VOC. Specifically, we identify 20 classes that are present in both two datasets and marked them as "novel classes". The other 60 classes that are unique to MS COCO are considered "base classes". To fine-tune our models, we use the subset of the MS COCO training set that contains only the base classes, while for testing, we use the validation set of PASCAL VOC that contains only the novel classes. This experimental setting is referred to as COCO2VOC.

    As illustrated in table~\ref{tab:fewshot}, we conduct experiments with different values of $N$ and $K$, and use $AP_{50}$ as the quantitative measure. Our UPLVP method gains a greater performance than the other few methods. Compared to FGN~\cite{fgn}, there are gains of \textbf{+2.1} $AP_{50}$ on $1-way~1-shot$ task and \textbf{+0.9} $AP_{50}$ on $3-way~3-shot$ task. Compared to vanilla K-Net~\cite{knet}, there are gains of \textbf{+2.5} $AP_{50}$ on $1-way~1-shot$ task and \textbf{+1.7} $AP_{50}$ on $3-way~3-shot$ task.

\begin{figure*}[t]
	\centering
	\subfloat[Pseudo masks in high quality.]{
        \includegraphics[width=0.97\linewidth]{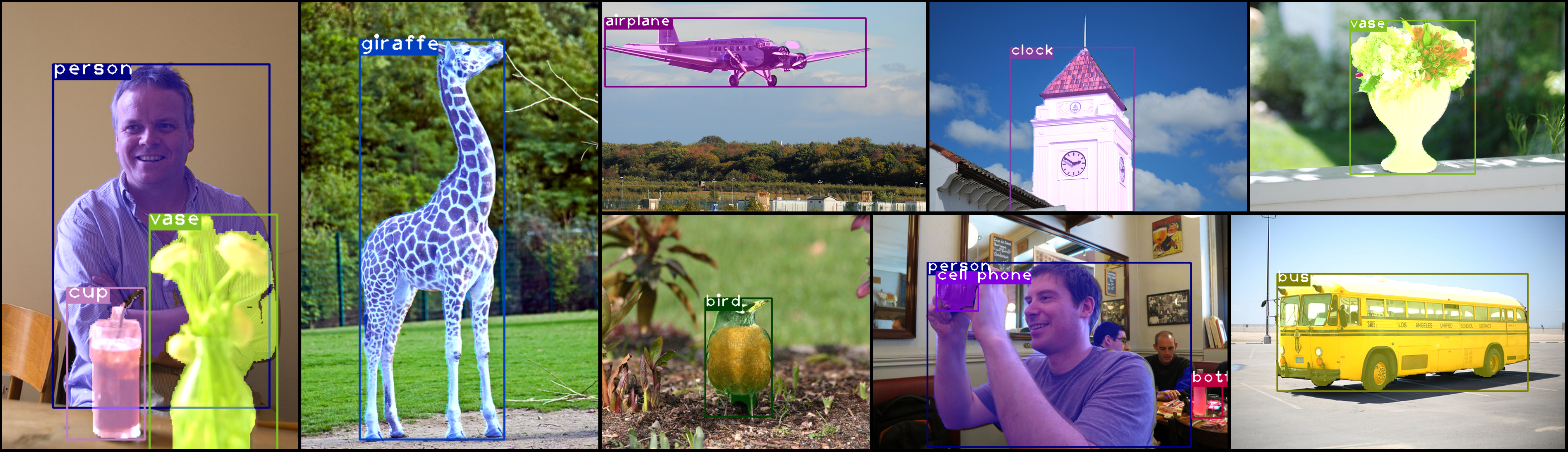}
        \label{masks-good}
        }

    \subfloat[Failure cases containing pixel noise or surrounding objects.]{
        \includegraphics[width=0.97\linewidth]{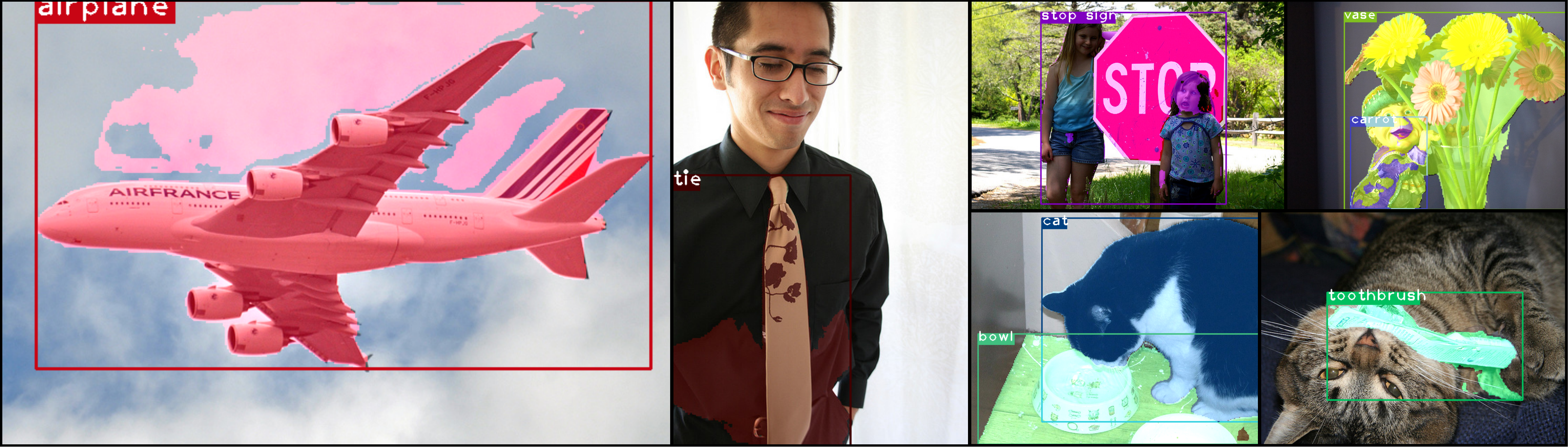}
        \label{pixel-noise}
        }

    \subfloat[Failure cases predicting incorrect categories.]{
        \includegraphics[width=0.97\linewidth]{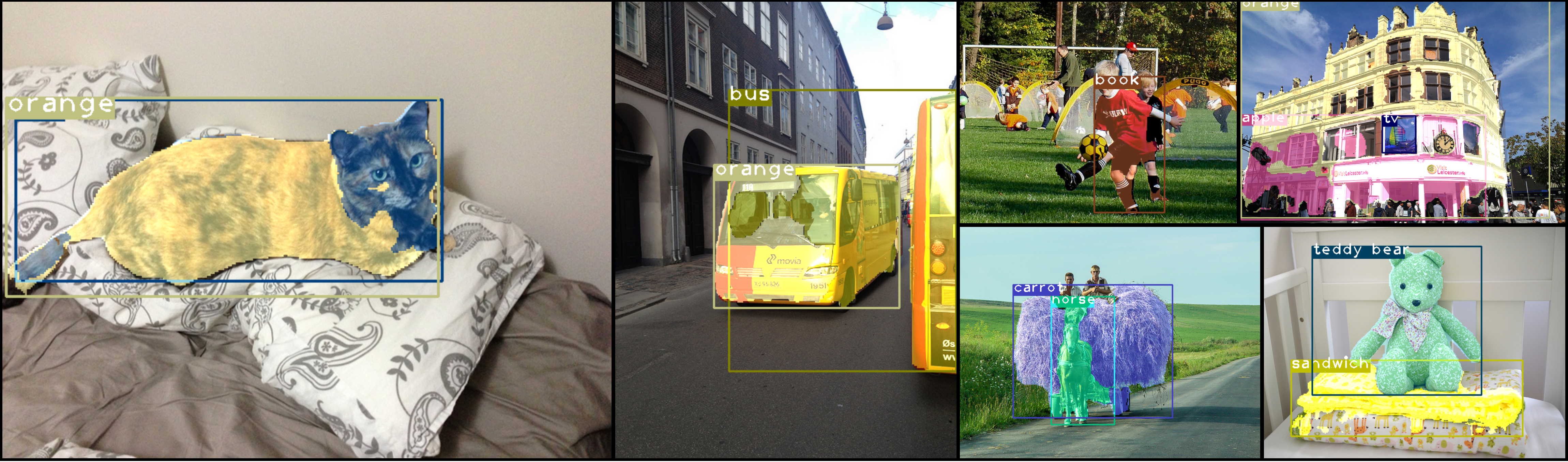}
        \label{wrong-cls}
        }
	\centering
	\caption{Visualization of pseudo masks with different quality. Although many high-quality masks exist (top), failure cases still occur. We divide the failure cases into two types: failure cases containing pixel noise or surrounding objects (middle) and failure cases predicting incorrect categories (bottom).}
	\label{psuedo-masks}
\end{figure*}

\subsection{Ablation Study}
    In this section, we conduct ablation experiments and analyze the results to fully understand the influence of each component of our UPLVP pre-train models.
    In general, we use K-Net as the QEIS model and pre-train in our method on COCO unlabeled2017 for 12 epochs and then fine-tune on COCO train2017 with 10\% images for 48 epochs.

\begin{figure*}[t]
  \centering
  \includegraphics[width=0.97\linewidth]{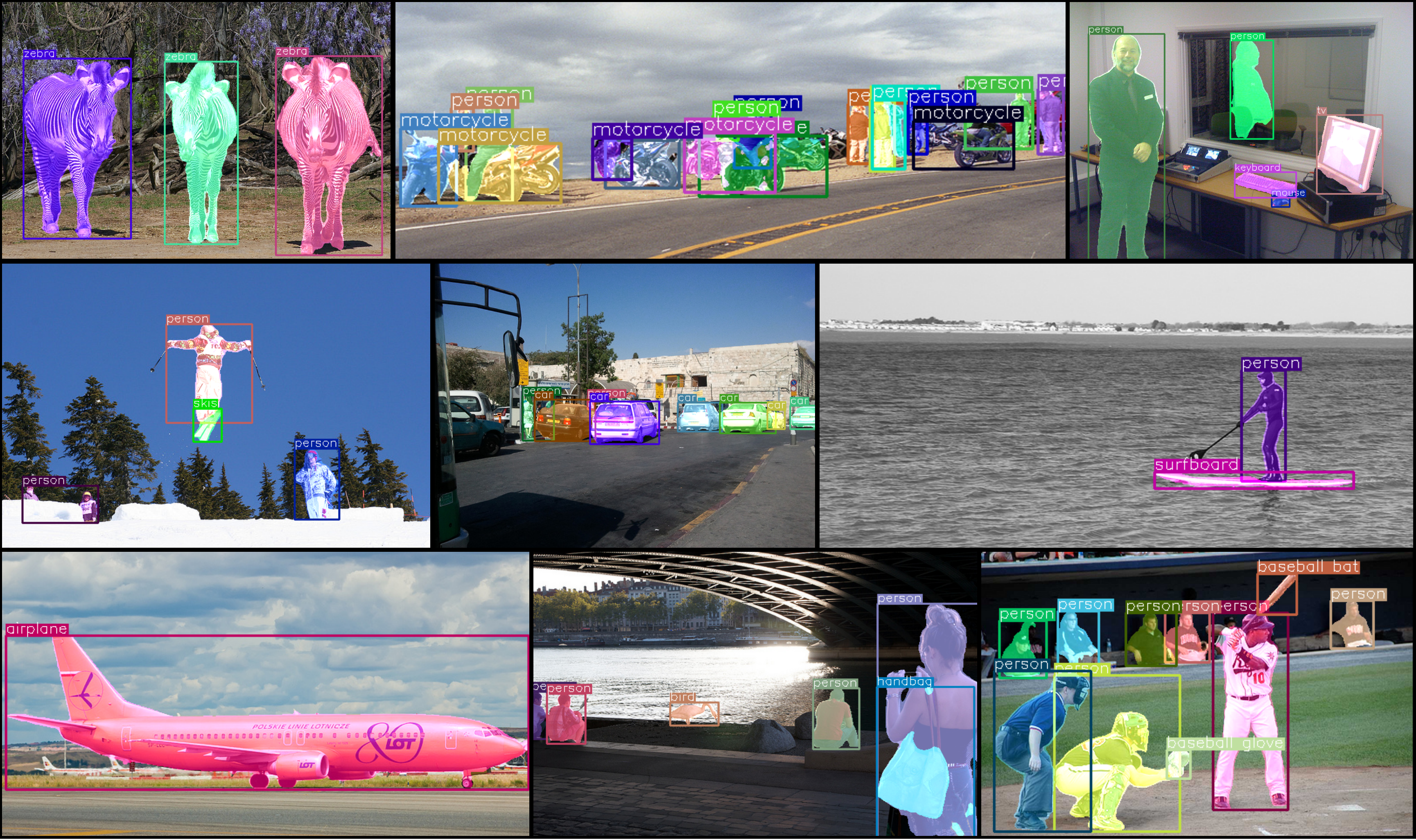}
  \caption{Visualization of prediction results using K-Net which is pre-trained on COCO train2017 and fine-tuned on COCO 10\% images. Each instance is distinguished by a distinct color.}
  \label{fig:psuedo-label}
\end{figure*}


\begin{figure*}[ht]
	\centering
	\includegraphics[width=1\linewidth]{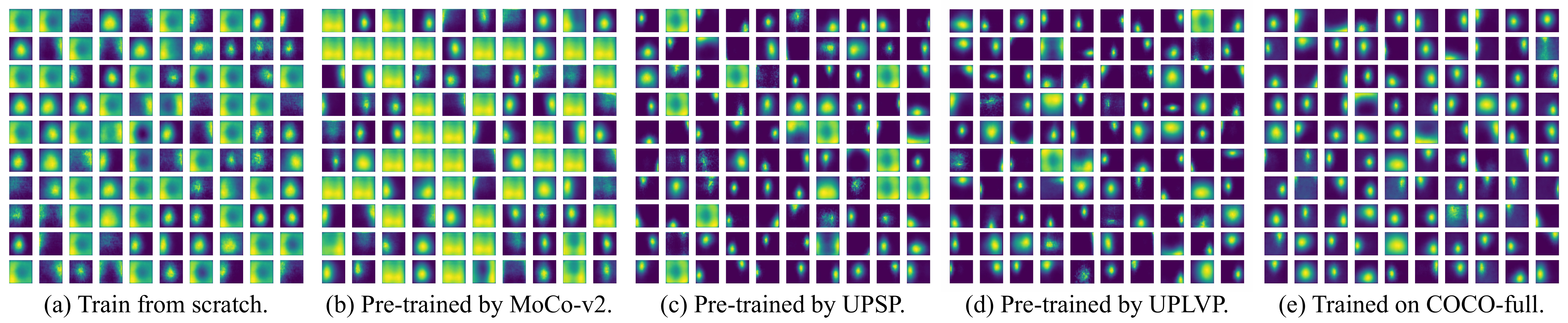}
	\caption{Average activation of the 100 kernels over 5000 images on COCO val2017. All masks are resized to $200\times 200$ for analysis. For each kernel, the highlighted region represents the main focus area. We can observe that in low data regimes, kernels trained from scratch (a) and pre-trained by MoCo-v2 (b) concentrate on redundant regions and shapes, making too many similar highlights among all kernels and some kernels directly concentrate on the whole image. Our prior work \cite{upsp} addressed this issue by introducing the UPSP and obtained much diverse highlight region distribution, while in the mean time, the UPLVP framework proposed in this paper (d) can further make each of the highlight regions more compact and the overall kernel region distribution more close to those trained with larger amounts of data (e). }
	\label{fig:visualize}
\end{figure*}

\subsubsection{Loss function} 
\label{abl:UPSP-loss}
    In this study, we have analyzed the effectiveness of the auxiliary kernel supervision loss. The results displayed in table~\ref{tab:loss_sem-ablation} demonstrate that this loss function has a positive impact on the mean average precision (mAP) by improving it by \textbf{+0.1}. This suggests that the auxiliary kernel supervision complements the prediction supervision. The analysis also shows that for large-scale instances, there is a significant increase of \textbf{+0.4} $AP_L$, while for smaller instances, there is a slight decrease of \textbf{-0.2} $AP_S$. This indicates that the auxiliary kernel loss is more beneficial for detecting large-scale objects that contain more pixels.

\subsubsection{Cosine Similarity Based Matching} 
\label{abl:UPSP-cosine}
    Table~\ref{tab:prompt-ablation} presents the evaluation results of several different prompt-kernel matching approaches, containing \emph{No Prompting, Random Assignment, Sequential Assignment and Cosine Similarity} based matching. 
    In the \emph{No Prompting} experiment, compared to the \emph{Cosine Similarity} matching, there is an increase by \textbf{+0.2} mAP.
    As for \emph{Random Assignment}, also named 'shuffle' in UP-DETR~\cite{updetr}, leading to a performance drop of \textbf{-0.1} mAP compared to the method even without using prompt. It is intuitively believed that shuffling the prompts misleads the matching between prompts and kernels, and thus prompts may be injected into distinct kernels.
    \emph{Sequential Assignment} simply expands the number of language-vision prompts to the number of kernels(set to 100 in our work) and attaches them to the initial kernels of K-Net, which achieves \textbf{+0.2} mAP improvement compared to Random Assignment. 
    Subsequently, \emph{Cosine Similarity} further surpasses Sequential Assignment by \textbf{+0.1} mAP and is also the best result among all prompt-kernel matching methods. The above experiments show that using cosine similarity matching helps to find the best matching kernels.

\subsubsection{Class-agnostic Object Detection}
\label{abl:UPSP-detec}
    We conduct the same operation as FreeSOLO~\cite{wang2022freesolo}, namely converting our proposed pseudo masks into detection bounding boxes. The results are compared with UP-DETR~\cite{updetr} and DETReg~\cite{detreg}. We experiment with each method with the COCO val2017 dataset and calculate the AP of their class-agnostic object detection results, as shown in table~\ref{tab:class_agnostic}.
    As can be seen, compared to the other two pre-training methods without self-training, UP-DETR and DETReg, our method achieves better performance and gains a 6.0 mAP value.
    With the help of two language-vision models, the mAP of our pseudo masks even surpasses the one gained by FreeSOLO, which requires multiple self-training steps and calls for much longer training time (extra 14 hours) and larger memory cost. 
    
    \noindent \textbf{Mask Quality Analysis.}
    In Table~\ref{tab:class_agnostic}, we can observe that the pseudo masks generated by two language-vision models have better quality and achieve an increase of \textbf{+0.5} mAP when compared with FreeSOLO, which requires a self-training process. However, we have noticed that although UPLVP has the highest $AP_{75}$ among UPLVP, FreeSOLO and UPSP, it has the lowest $AP_{50}$. This indicates that both good and bad masks increase in number. In this subsection, we showcase pseudo masks of different qualities and present them visually in Figure~\ref{psuedo-masks}.
    
    The best pseudo masks are shown in Figure~\ref{masks-good} which are able to precisely segment objects and predict categories comparatively. Moreover, they also have minimal pixel noise outside the objects, allowing for tight bounding boxes. 
    However, there are many pseudo masks that are of low quality, which can be divided into two types, \emph{failure masks containing pixel noise or surrounding objects} and \emph{failure masks predicting incorrect categories}. 
    The first type, shown in Figure~\ref{pixel-noise}, includes masks that contain pixel noise or pixels from surrounding objects(the sky behind the plane), even though they can segment the whole object(the plane). This might be due to the fact that CLIP tends to give more attention to contextual words~\cite{textclip}. 
    The second type, shown in Figure~\ref{wrong-cls}, includes masks that may accurately segment objects but fail to predict their correct categories. These two types of problems can negatively impact the quality of pseudo masks and pre-training performance. To address these issues, we plan to optimize the quality of pseudo masks in the future.

\begin{table}[t]
\renewcommand{\arraystretch}{1.2}
\caption{Ablation of different prompt approaches. '\XSolidBrush' means the model only pre-trained by pseudo labels without prompt. 'Cos. Sim.' means Cosine Similarity Assignment.}\vspace{-8pt}
\centering
\begin{tabular}{ccccccc}
\hline
Prompt method  & mAP  & $AP_{50}$ & $AP_{75}$ & $AP_{S}$ & $AP_{M}$ & $AP_{L}$ \\
\hline
\XSolidBrush      & 25.2 & 44.1   & 25.1   & 8.6   & 26.7  & 40.9  \\
Rand. Assign     & 25.2 & 43.9   & 25.2   & 8.3   & 27.1  & 41.5  \\
Seq. Assign   & 25.3   & 44.2   & 25.7     & \textbf{8.9}     & 27.0  & 41.9  \\
\rowcolor[HTML]{EFEFEF} 
\textbf{Cos. Sim.} & \textbf{25.4} & \textbf{44.4} & \textbf{25.7} & 8.8 & \textbf{27.2} & \textbf{42.0}  \\\hline
\end{tabular}
\label{tab:prompt-ablation}
\vspace{-10pt}
\end{table}

\begin{table}[t]
\renewcommand{\arraystretch}{1.2}
\caption{Unsupervised class-agnostic object detection results. Here we evaluate the AP of pseudo masks generated by different methods. '\XSolidBrush' refers to a technique that is entirely unsupervised and does not require self-training."}\vspace{-8pt}
\centering
\footnotesize
\begin{tabular}{ccccc}
\hline
Method       & Self-train?                                & AP  & $AP_{50}$ & $AP_{75}$ \\ \hline
FreeSOLO~\cite{wang2022freesolo}     & {\color[HTML]{009901} \Checkmark}          & 5.5 & \textbf{12.2}    & 4.2     \\
UP-DETR~\cite{updetr}      & {\color[HTML]{CB0000} \XSolidBrush}        & 0   & 0       & 0       \\
DETReg~\cite{detreg}       & {\color[HTML]{CB0001} \XSolidBrush}        & 1.0   & 3.1     & 0.6     \\
UPSP~\cite{upsp}             & {\color[HTML]{CB0002} \XSolidBrush}        & 3.2 & 8.5     & 2.0       \\
\rowcolor[HTML]{EFEFEF} 
UPLVP & {\color[HTML]{CB0002} \XSolidBrush}        & \textbf{6}   & 8.4    & \textbf{5.6}       \\ \hline
\end{tabular}
\label{tab:class_agnostic}
\vspace{-10pt}
\end{table}

\begin{table}[t]
\renewcommand{\arraystretch}{1.2}
\centering
\caption{Ablation of Pseudo Labels on COCO with 10\% images. P means our Prompting method.} \vspace{-8pt}
\begin{tabular}{ccccc}
\hline
\rowcolor[HTML]{FFFFFF} 
\multicolumn{2}{l}{\cellcolor[HTML]{FFFFFF}Approch}                   & \multicolumn{2}{l}{\cellcolor[HTML]{FFFFFF}Pseudo Label (quality)} & mAP  \\ \hline
\multicolumn{2}{l}{K-Net $w/o$ P}                       & \multicolumn{2}{l}{Rand. Prop. (bad)}                               & 0.8 \\
\multicolumn{2}{l}{K-Net $w/$ P}                         & \multicolumn{2}{l}{Rand. Prop. (bad)}                               & 10.0 \\ 
\multicolumn{2}{l}{K-Net $w/o$ P}                       & \multicolumn{2}{l}{UPSP \color[HTML]{CB0000}{(normal)}}                              & 21.7\\
\multicolumn{2}{l}{K-Net $w$ P}    & \multicolumn{2}{l}{{UPSP \color[HTML]{CB0000}{(normal)}}}      & 23.5 \\
\multicolumn{2}{l}{K-Net {$w/o$ P}}  & \multicolumn{2}{l}{FreeSOLO \color[HTML]{009901}{(good)}}      & 23.3 \\
\multicolumn{2}{l}{K-Net $w/$ P}     & \multicolumn{2}{l}{FreeSOLO \color[HTML]{009901}{(good)}}                               & 24.0 \\ 

\rowcolor[HTML]{EFEFEF} 
\multicolumn{2}{l}{\cellcolor[HTML]{EFEFEF}K-Net $w/o$ P}    & \multicolumn{2}{l}{\cellcolor[HTML]{EFEFEF}UPLVP \color[HTML]{009901}{(good)}}        & \textbf{25.2}\\
\rowcolor[HTML]{EFEFEF} 
\multicolumn{2}{l}{\cellcolor[HTML]{EFEFEF}K-Net $w/$ P}     & \multicolumn{2}{l}{\cellcolor[HTML]{EFEFEF}UPLVP \color[HTML]{009901}{(good)}}      & \textbf{25.4} \\ \hline

\end{tabular}
\label{tab:pseudo-label}
\vspace{-10pt}
\end{table}

\begin{table}[t]
\renewcommand{\arraystretch}{1.2}
\caption{Ablation of {80 COCO classes gained by CLIP}. 'Saliency' means we transfer the class labels to 0/1 representing background/foreground.} \vspace{-8pt}
\centering
\setlength{\tabcolsep}{1.7mm}{
\begin{tabular}{ccccccc}
\hline
 Approach           & mAP           & $AP_{50}$        & $AP_{75}$        & $AP_{S}$        & $AP_{M}$         & $AP_{L}$         \\ \hline
\rowcolor[HTML]{EFEFEF} 
  UPLVP (original) & \textbf{25.4} & \textbf{44.4} & \textbf{25.7} & \textbf{8.8} & \textbf{27.2} & \textbf{42.0} \\
  UPLVP (saliency)      & 24.2          & 42.4          & 24.5          & 8.2          & 26.0          & 39.0       \\ 
   UPSP~\cite{upsp} & 23.5 & 41.4 & 23.7 & 7.9 & 24.8 & 38.6 \\ \hline
\end{tabular}}
\label{tab:80categories-ablation}
\vspace{-10pt}
\end{table}

\subsubsection{Pseudo Mask Analysis} 
\label{abl:UPSP-mask}
    In this section, we experiment and analyze our prompting method on several kinds of pseudo labels in different qualities. 
    As illustrated in table~\ref{tab:pseudo-label}, pseudo labels in the worst quality generated randomly(\emph{Random Prompt}) have a negative impact on the results of the fine-tuned model since randomly generated prompts destroy the localization and shape priors of kernels to some extent.
    However, our pre-training method approaches significant improvement by utilizing language-vision prompts no matter whether kernels are injected by prompts. Besides, our method can also outperform UPSP~\cite{upsp} which lacks category information, and FreeSOLO~\cite{wang2022freesolo} which requires a much longer training time and many process steps.
    Moreover, our Prompting method achieves competitive performance with both normal and good-quality pseudo masks. With our method, UPSP, FreeSOLO and our UPLVP obtain an increase by \textbf{+1.8}, \textbf{+0.7} and \textbf{+0.2} mAP, respectively. 
    
    Except for the quality of pseudo labels, category information brought by CLIP is crucial as well. For UPSP and FreeSOLO, pseudo masks are generated in a saliency mechanism thus containing only foreground and background. Here we explore the contribution of the category information by following the saliency mechanism and simply convert the category labels to \{0,1\} and the 0/1 represents background/foreground. As shown in table~\ref{tab:80categories-ablation}, comparing to the UPLVP with saliency mechanism and UPSP~\cite{upsp}, our method gains an increase by \textbf{+1.2} and \textbf{+1.9} mAP respectively, indicating that labels with 80 categories are essential to our pre-training model, verifying category information that brought by language-vision models has the capability to promote the learning of models.
    We further visualize our prediction fine-tuned with 10\% COCO images in Figure~\ref{fig:psuedo-label}. 

\subsection{Kernel Spatial Distribution Analysis}
    To verify whether the proposed language-vision prompts can inject localization and shape priors into the kernels/queries of prediction heads in QEIS models, we further visualize the average activation masks of 100 instance kernels in K-Net over 5000 images on the 2017val split after 2 training epochs. 

    As can be seen, the best result (Figure~\ref{fig:visualize}d) is from the fully trained K-Net on COCO train2017. With the whole COCO dataset, each instance kernel is trained to concentrate on a specific region with diverse locations and shapes. 
    This can also verify the great performance of QEIS models with a large amount of data. 
    However, the priors from the supervised (Figure~\ref{fig:visualize}a) and the compared unsupervised (Figure~\ref{fig:visualize}b) method mainly focus on the central area, many of which attend to a similar region lacking location and shape diversity. This experiment shows the weaknesses of QEIS models in low data regimes, and the pre-training methods that only adjust the backbones are not able to address the issue effectively.
    
    Surprisingly, the kernels learned through our unsupervised pre-training method (Figure~\ref{fig:visualize}e) present positive trends in the diversity of localization and shape priors, which can approach the fully trained kernels. Compared with the previous work UPSP~\cite{upsp} (Figure~\ref{fig:visualize}c), there is an obvious improvement in the kernels using our UPLVP which can focus on more precise locations.
    The above results demonstrate that the kernels pre-trained with language-vision prompts have learned effective spatial distribution and shape discrimination ability. With the help of kernels injected by language-vision prompts, QEIS models have the capacity to speed up convergence even with small scale datasets and reach a comparable performance.

\section{Conclusion}
    This paper first points out the shortcomings of QEIS models, namely lacking spatial distribution and shape awareness with small datasets and thus performing poorly in low-data regimes.
    Hence we propose \emph{Unsupervised Pre-training with Language-Vision Prompt}, a novel unsupervised pre-train method leveraging visual prompts gained from language-vision models, which can significantly boost the convergence speed and performance of QEIS models in low-data instance segmentation tasks and reach comparable or even better performance comparing to CNN-based models.
    From the perspective of technical, it is the first paper that explores the application of generating prompts by language-vision models and prompting in the field of instance segmentation. We hope that our novel design elements are able to provide innovative insights for works in the future on visual prompting mechanisms. In future work, we will follow more related studies on visual prompts to further promote the prompt learning mechanism and apply the prompt learning mechanism to advance the weakly supervised learning community~\cite{zhang2020weakly,zhang2021weakly,zhao2021weakly}. 

\noindent \textbf{Limitations.}
    The majority of the masks we have generated are large in size, which makes our pre-trained kernels/queries mostly concentrate on objects with more pixels rather than small objects. Compared to the predicting accuracy improvements on large instances, our pre-training method achieves limited improvement on small ones. The other is there is still room for improvement in the quality of our proposed masks, many of which tend to contain their surrounding context(like environment or crowded parts). Therefore, an appropriate method is required to refine the pseudo masks. We believe there is plenty of room to further optimize our proposed method.

\bibliographystyle{IEEEtran}
\bibliography{egbib}

\vspace{-0.5in}
\begin{IEEEbiography}[{\includegraphics[width=1in,height=1.25in,clip,keepaspectratio]{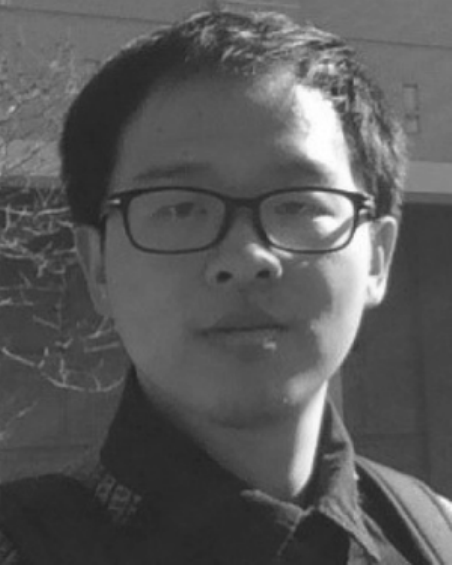}}]{Dingwen Zhang}    
    is a professor with School of Automation, Northwestern Polytechnical University, Xi'an, China. He received his Ph.D. degree from NPU in 2018. From 2015 to 2017. he was a visiting scholar at the Robotic Institute, Carnegie Mellon University, Pittsburgh, United States. His research interests include computer vision and multimedia processing, especially on saliency detection and weakly supervised learning.
\end{IEEEbiography}
\vspace{-0.45in}
\begin{IEEEbiography}[{\includegraphics[width=1in,height=1.25in,clip,keepaspectratio]{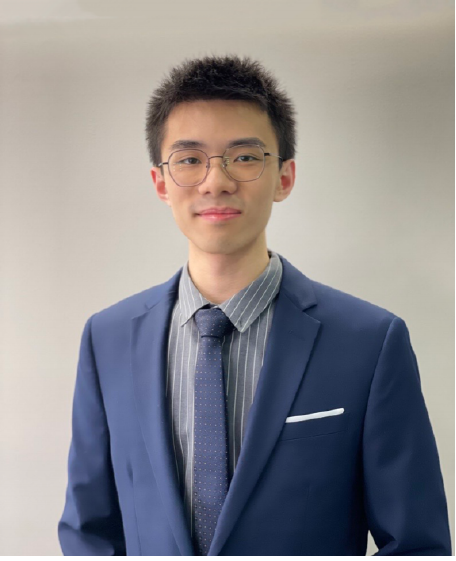}}]{Hao Li}    
     received the B.E. degree from Beijing University of Chemical Technology, Beijing, China, in 2022. He is currently working toward the Ph.D. degree in the School of Automation, Northwestern Polytechnical University, Xi’an, China. His research interests include computer vision and pattern recognition, especially on weakly supervised object segmentation, instance segmentation and 3D scene understanding.
\end{IEEEbiography}
\vspace{-0.55in}
\begin{IEEEbiography}[{\includegraphics[width=1in,height=1.25in,clip,keepaspectratio]{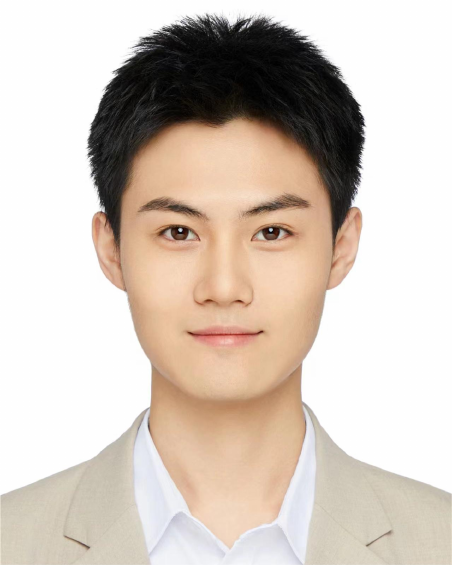}}]{Diqi He}    
    received the B.E. degree from Northwestern Polytechnical University, Xi'an, China. He is currently working toward an M.E. degree in the School of Automation, Northwestern Polytechnical University, Xi’an, China. His research interests include computer vision and pattern recognition, especially on vision segmentation and large scale models.
\end{IEEEbiography}
\begin{IEEEbiography}[{\includegraphics[width=1in,height=1.25in,clip,keepaspectratio]{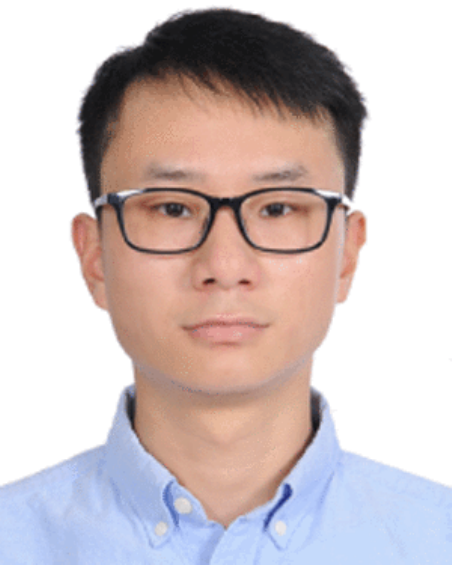}}]{Nian Liu}    
     received the BS degree and the PhD degree from the School of Automation, Northwestern Polytechnical University, Xi’an, China, in 2020 and 2012, respectively. He is currently a research scientist with Mohamed Bin Zayed University of Artificial Intelligence, Abu Dhabi, UAE. His research interests include computer vision and deep learning, especially on saliency detection and few shot learning.
\end{IEEEbiography}
\vspace{-3.5in}
\begin{IEEEbiography}[{\includegraphics[width=1in,height=1.25in,clip,keepaspectratio]{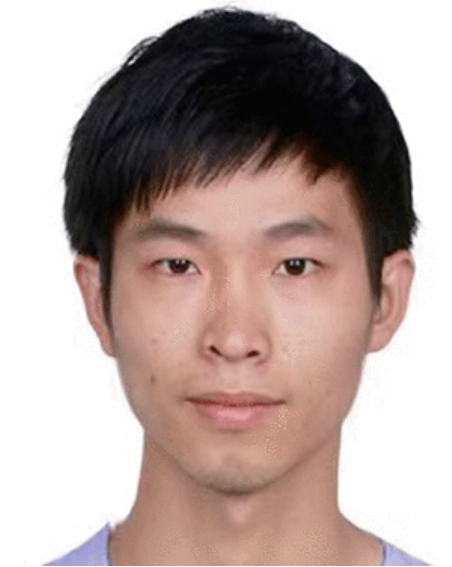}}]{Lechao Cheng}    
     (Member, IEEE) received the Ph.D. degree from the College of Computer Science and Technology at Zhejiang University, Hangzhou, China, in 2019. He is currently an Associate Professor at the School of Computer Science and Information Engineering, Hefei University of Technology, Hefei, China. He has contributed more than 40 research papers to renowned academic journals and conferences such as IJCV, TMI, TCyb, TMM, CVPR, AAAI, IJCAI, and ACM MM. His research area centers around vision knowledge transfer in deep learning.
\end{IEEEbiography}
\vspace{-3.5in}
\begin{IEEEbiography}[{\includegraphics[width=1in,height=1.25in,clip,keepaspectratio]{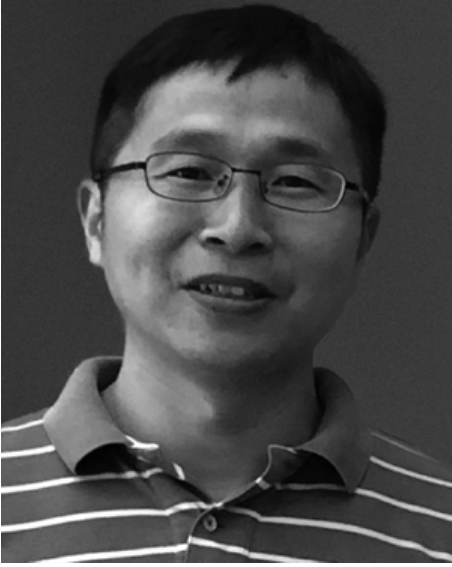}}]{Jingdong Wang}    
     received the BEng and MEng degrees from the Department of Automation, Tsinghua University, Beijing, China, in 2001 and 2004, respectively, and the PhD degree from the Department of Computer Science and Engineering, The Hong Kong University of Science and Technology, Hong Kong, in 2007. He is currently a senior principal research manager at the Visual Computing Group, Microsoft Research, Beijing, China. His research interests include neural network design, human pose estimation, semantic segmentation, large-scale indexing, and person re-identification. He is an associate editor of the IEEE Transactions on Pattern Analysis and Machine Intelligence and IEEE Transactions on Circuits and Systems for Video Technology, and is also an area chair of several leading computer vision and AI conferences, such as CVPR, ICCV, ECCV, ACM MM, IJCAI, and AAAI. He is an IAPR fellow and an ACM distinguished member.
\end{IEEEbiography}
\vspace{-3.5in}
\begin{IEEEbiography}[{\includegraphics[width=1in,height=1.25in,clip,keepaspectratio]{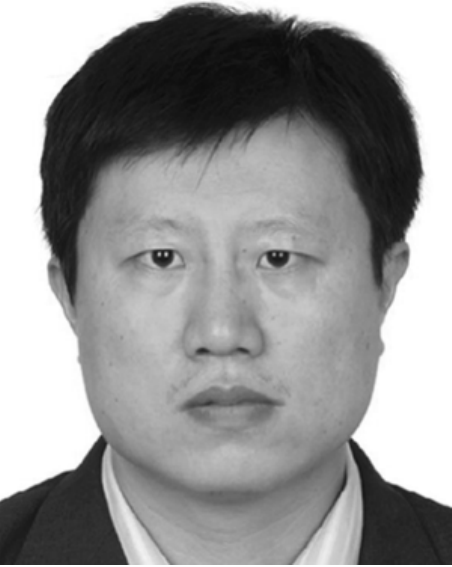}}]{Junwei Han}    
    (M’12-SM’15) received the PhD degree from Northwestern Polytechnical University, in 2003. He is a professor with Northwestern Polytechnical University, Xi’an, China. His research interests include computer vision and brain imaging analysis. He has published more than 100 papers in IEEE TRANSACTIONS and top tier conferences. He is a senior member of the IEEE.
\end{IEEEbiography}

\end{document}